\documentclass{article}

\usepackage{PRIMEarxiv}

\usepackage[utf8]{inputenc} 
\usepackage[T1]{fontenc}    
\usepackage{hyperref}       
\usepackage{url}            
\usepackage{booktabs}       
\usepackage{amsfonts}       
\usepackage{nicefrac}       
\usepackage{microtype}      
\usepackage{lipsum}
\usepackage[utf8]{inputenc}
\usepackage{fancyhdr}       
\usepackage{graphicx}       
\graphicspath{{media/}}     
\usepackage{multirow}
\usepackage[most]{tcolorbox}
\title{PersonaBOT: Bringing Customer Personas to Life with LLMs and RAG
\thanks{This study was carried out as part of a Master's thesis project at Volvo Construction Equipment.}
}

\author{
  Muhammed Rizwan\textsuperscript{1, 2}, Lars Carlsson\textsuperscript{1}, Mohammad Loni\textsuperscript{2, $\dagger$}
 \\
  \\\textsuperscript{1}Department of Computer Science, Jönköping University, Jönköping, Sweden \\
  \textsuperscript{2}Department of Future Solutions, Volvo Construction Equipment, Eskilstuna, Sweden \\
  \textsuperscript{$\dagger$}\textit{Corresponding author}: \texttt{mohammad.loni@volvo.com}
}

\begin{document}
\maketitle

\begin{abstract}

The introduction of Large Language Models (LLMs) has significantly transformed Natural Language Processing (NLP) applications by enabling more advanced analysis of customer personas. At Volvo Construction Equipment (VCE), customer personas have traditionally been developed through qualitative methods, which are time-consuming and lack scalability. The main objective of this paper is to generate synthetic customer personas and integrate them into a Retrieval-Augmented Generation (RAG) chatbot to support decision-making in business processes. To this end, we first focus on developing a persona-based RAG chatbot integrated with verified personas. Next, synthetic personas are generated using Few-Shot and Chain-of-Thought (CoT) prompting techniques and evaluated based on completeness, relevance, and consistency using McNemar's test. In the final step, the chatbot’s knowledge base is augmented with synthetic personas and additional segment information to assess improvements in response accuracy and practical utility. Key findings indicate that Few-Shot prompting outperformed CoT in generating more complete personas, while CoT demonstrated greater efficiency in terms of response time and token usage. After augmenting the knowledge base, the average accuracy rating of the chatbot increased from 5.88 to 6.42 on a 10-point scale, and 81.82\% of participants found the updated system useful in business contexts.

\end{abstract}

\keywords{Customer Persona \and Market Analysis \and LLM \and RAG}

\section{Introduction}
\label{sec:introduction}

The advent of large language models (LLMs) has significantly advanced the field of natural language processing (NLP). These models are capable of capturing complex linguistic patterns and are increasingly employed in diverse applications, including virtual assistance, text generation \cite{loni2024review}, and information extraction \cite{hadillms}. Their adoption across industries has enabled new levels of automation and insight, particularly in customer-facing domains such as marketing, customer support, and strategic planning \cite{cheung2024reality}.

Customer personas—detailed representations of user segments—play a pivotal role in enabling businesses to tailor their offerings and communication strategies. Traditionally, personas are crafted through qualitative methods such as interviews and surveys, which, while insightful, are time-intensive and difficult to scale \cite{sun2024persona}. More recently, data-driven approaches have emerged, leveraging statistical and machine learning techniques to streamline persona creation \cite{mcginn2008data,zhang2016data,jung2018automatic}. However, these methods often struggle to extract nuanced insights from unstructured text and adapt to evolving customer behavior in real time.

LLMs such as GPT-4 offer a new avenue for generating high-quality, structured customer personas directly from unstructured textual data, such as customer success stories \cite{de2023improved, salminen2024deus, sun2024persona}. Nevertheless, current research has primarily focused on using a single prompting method, without comparing the effectiveness of alternative strategies like few-shot or chain-of-thought (CoT) prompting \cite{wei2022chain, brown2020language}. Moreover, the practical integration of generated personas into business workflows—especially through interactive systems like retrieval-augmented generation (RAG) chatbots—remains underexplored \cite{lewis2020retrieval}.

This paper addresses these gaps by: (1) evaluating the effectiveness of different prompting techniques for generating synthetic customer personas from publicly available texts; and (2) presenting a proof-of-concept chatbot system that enables users to interact with these personas through natural language queries. The study focuses on a use case from the construction industry, where customer segmentation is critical but existing persona development practices are resource-intensive and static.

By exploring both the generation and application of synthetic personas, this work contributes to the growing body of research at the intersection of LLMs, human-centered design, and business decision support. It provides practical insights into how organizations can adopt LLM-based tools for scalable, data-driven persona generation and utilization.

\section{Related Work}
\label{sec:relatedworks}

\subsection{Non-LLM Approaches for Creating Customer Personas}
\label{section3.1}

The traditional method of creating personas depended on qualitative data, such as interviews, observations, and survey data from target users \cite{sun2024persona}. There have been researches that explored data-driven approaches that improved efficiency, scalability, and reliability in creating personas. One such approach was introduced by McGinn et al. \cite{mcginn2008data}, where a survey was sent over to 1300 users. An exploratory factor analysis, a data reduction technique, was performed on the survey results. This analysis helped identify the groups based on the tasks performed. Stakeholders were involved throughout this process to ensure the relevance of personas. Instead of relying on survey data or user interviews, Zhang et al.\cite{zhang2016data} followed a two-step statistical machine-learning approach to create personas only based on user behavior. In the first step, they analyzed 3.5 million clicks from 2400 users and clustered them into a common workflow using hierarchical clustering. In the second step, a mixed statistical model was used to create five personas. 

Jung et al.\cite{jung2018automatic} introduced Automatic Persona Generation (APG), a system that creates personas from real-time social media interactions on platforms like Facebook and YouTube. They processed tens of millions of interactions using non-negative matrix factorization. This automatically generated realistic and up-to-date personas from large-scale social media data. Similarly, Farseev et al.\cite{farseev2024somonitor} introduced a framework named SOMONITOR that used X-Mean clustering with ADA embeddings to extract customer personas from digital marketing content. Unlike previous studies that relied on survey data or behavioral data, SOMONITOR clusters advertising content into distinct persona groups based on customer needs, interests, and aspirations. While these data-driven methods significantly improve persona creation, advancements in LLM offer further opportunities for automating and improving persona development.

\subsection{LLM Approaches of Creating Personas}
\label{section3.2}

LLMs' ability to generate structured text based on the input text provided using advanced natural language processing capabilities makes them a strong candidate for persona creation. This section reviews various approaches that utilized LLMs for persona creation.

One of the methods used to create personas is by providing LLM with structured prompts. This methodology was used in \cite{salminen2024deus} to create 450 personas. In this study, they utilized these generated personas and investigated the bias and diversity in them. Their findings indicated that LLMs can create informative and relatable personas, but they exhibit a strong bias from specific countries. Similarly, Zhang et al. \cite{zhang2023personagen} introduced PersonaGen, a tool that used Generative Pre-trained Transformer (GPT)-4 \cite{achiam2023gpt} along with knowledge graphs to refine persona generation. The tool was developed to assist the agile software development process. The GPT-4 model analyzed the user feedback provided and generated high-quality and detailed persona content. This content was then used by the knowledge graphs to create personas. PersonaGen demonstrated that it improved accuracy in capturing user needs compared to independent human analysis. Although challenges remain in analyzing non-functional requirements.

Another method for persona generation using LLMs involves the use of thematic analysis. De Paoli et al. \cite{de2023improved} proposed a workflow where LLMs analyze qualitative interview data to generate personas. This approach follows a structured methodology where LLMs first generate codes (such as behaviors, goals, etc.) in textual format. From these codes, emerging themes are identified. These themes, along with prompts, are then used to construct persona narratives. The advantage of this method lies in its ability to extract meaningful user traits from raw interview data without predefined coding schemes. An extension of this approach is found in Persona-L \cite{sun2024persona}, a system that integrates LLMs with a RAG framework. By using specific types of datasets, this system enhances persona realism while addressing biases commonly found in LLM-generated content. The system was tested in creating personas that represent individuals with complex needs. This study demonstrated that incorporating external data can improve both the diversity and contextual accuracy of the generated personas.

Beyond structured prompting and thematic analysis, there has been a study that used  human-AI collaboration in persona generation. Goel et al. \cite{goel2023preparing} conducted an exploratory study where novice designers used GPT-3 \cite{brown2020language} to create personas through iterative refinement. The study found that personas generated with GPT-3 were comparable to those created manually, particularly when designers provided detailed prompts and engaged in multiple iterations. However, the study also highlighted challenges such as generic responses, inconsistencies, and stereotypical outputs. This makes it necessary for human intervention to refine and personalize the generated personas.

\subsection{Non-LLM Approaches for Analyzing and Leveraging Customer Personas}
\label{section3.3}

There have been various techniques to analyze and utilize customer persona before the emergence of LLMs. These methods were based on statistical methods \cite{morande2023digital} and machine learning \cite{more2025quantum,praveen2024crafting} to extract insights from customer data. This section reviews these approaches, limitations, and the reasons for the shift towards using LLM-based methods.

One such approach is the use of Quantum artificial intelligence (QAI). QAI combines quantum computing and AI to process large datasets in parallel. This allows obtaining real-time updates of customer profiles in response to dynamic behaviors and preferences. A study by More et al. \cite{more2025quantum} discussed how QAI can improve sentiment analysis and predictive modeling using quantum machine learning. This methodology improves customer segmentation, recommendation engines, and consumer behavior prediction. Even though QAI is promising, it remains in the early stages of adoption, and its implementation is challenging due to limited computational feasibility and hardware availability.

Generative AI techniques, such as Generative Adversarial Nets (GANs) \cite{goodfellow2014generative} and Variational Autoencoders (VAEs) \cite{kingma2013auto}, have been explored for improving marketing applications using personas. \cite{morande2023digital} demonstrated that GANs and VAEs can improve customer profiling in social media marketing by generating personalized product recommendations and marketing content. The study found that generated content was able to significantly improve customer engagement, loyalty, and sales. However, generative models cause risks related to algorithmic bias, ethical concerns about privacy, and the chances of misleading or made-up customer insights \cite{morande2023digital}.

Another approach includes utilizing Bayesian probabilistic models such as Latent Dirichlet Allocation(LDA) \cite{blei2003latent} and Structural Topic Models (STM) \cite{roberts2014structural}. These models categorize textual data into topics based on word co-occurrence patterns, helping in customer segmentation and persona identification. However, Bayesian models are frequency-based models that rely on word frequency distribution. They struggle to capture the complicated and nuanced elements in the textual data. This is due to the lack of an attention mechanism, a key feature of modern transformer-based LLMs \cite{praveen2024crafting}. The challenges discussed above in \cite{more2025quantum, morande2023digital, praveen2024crafting} have led researchers and businesses to adopt LLMs, as these models show promise in contextual understanding and adaptability.

\subsection{LLM Approaches for Analyzing and Leveraging Customer Personas}
\label{section3.4}

This section reviews studies that utilized LLMs for personas, such as (i) persona interpretation, (ii) personalization, (iii) role-playing techniques, (iv) investigating bias and stereotypes, and (v) business insights.

\textbf{(i) Persona Interpretation:} LLMs are built upon large and diverse datasets, enabling them to interpret and generate user personas with high precision. Unlike traditional methods that need structured datasets and predefined heuristics, LLMs can extract persona-related attributes from conversational data, social media posts, and customer feedback. This information can then be used to understand the needs, motivations, and goals of the specific user. \cite{panda2024llms} examines how LLM interprets culturally specific personas, focusing on the Indian context. The research conducted both quantitative and qualitative analyses to assess how well LLMs understood personas within cultural contexts. The study revealed that LLMs exhibit high consistency and completeness in persona evaluation, but they struggle with credibility. 

\textbf{(ii) Personalization}: One of the notable advancements in LLM-driven persona development is personalization. Zhang et al. \cite{zhang2024personalization} provides a detailed survey of how LLMs can be personalized. They propose a taxonomy of personalization levels in three categories: user-level, persona-level, and global preferences. This study highlights how techniques like RAG and prompt engineering can be used to tailor responses to user-specific needs.

There have been studies that have explored how LLMs can be customized for personalized interaction. One such application is CloChat \cite{ha2024clochat}, which allows users to tailor personas for various contexts and tasks. The end user of this application can choose to define personas by altering attributes like conversational style, emotions, areas of interest, and visual representations, thereby making interactions more human-like and relevant. To assess CloChat's effectiveness, researchers conducted surveys and in-depth interviews by comparing it with ChatGPT. The findings indicated that CloChat significantly improves user engagement, trust, and emotional connection when compared with ChatGPT. 

\textbf{(iii) LLM Role-playing}: Personas can be integrated with LLMs through two approaches: LLM Role-Playing and LLM Personalization. In LLM role-playing, LLMs are assigned personas (roles) and they adapt to specific environments and tasks. Whereas in LLM personalization, LLM is adapted to user-specific personas for customized responses. The techniques used in Role-Playing are prompt engineering, multi-agent frameworks, and emergent behaviors in specific domains. In the personalization techniques, user data is integrated by Reinforcement Learning from Human Feedback (RLHF), fine-tuning, and memory mechanisms. This study highlights several challenges associated with role-playing personas, including limited contextual understanding, the need for manual persona creation, and the static nature of personas, which prevents them from adapting to dynamic tasks \cite{tseng2024two}. To address these challenges, \cite{schreiber2024pattern} proposes a pattern language for persona-based interactions. This pattern language contains a series of patterns, where each pattern identifies a specific problem and provides its associated solution in the form of a template. In this study, seven person-related patterns are introduced, which improved realism, adaptability, and specificity in LLM interactions, making them more effective for complex and evolving tasks. 

\textbf{(iv) Investigate bias and stereotypes:}
While persona-based LLMs improve customer engagement, they also can introduce biases and stereotypes which may affect customer insights and segmentations. Cheng et al. \cite{cheng2023marked} introduced Marked Persona, a prompt-based framework that captures the patterns and stereotypes across the LLM outputs. Their study used GPT-3.5 and GPT-4 to generate personas across various demographic groups and analyzed how this output is different from human-written personas. The findings reveal that personas generated by LLM contain more stereotypes than the personas written by humans. These biases are a challenge for LLM driven customer analysis, as they can provide inaccurate customer information. 

\textbf{(v) Business insights:} Understanding customer preferences and requirements has become important for businesses.Extracting and analyzing customer data manually is often a difficult and time-consuming task. Barandoni et al. \cite{barandoni2024automating} evaluate the ability of proprietary and open-source models, such as GPT-4, Gemini, and Mistral 7B, to extract customer needs from TripAdvisor forum posts. This study systematically compared two prompting techniques, such as CoT using various proprietary and open-source LLMs for customer needs extraction. However, the focus was on extracting short customer needs from forum posts, not on generating structured personas. Additionally, the study did not explore fine-tuning techniques, which could have further improved the model's performance. In contrast, \cite{praveen2024crafting} used fine-tuning on different models to identify topics, emotions, and sentiments from TripAdvisor customer reviews. This study provides an alternative technique for improving LLM-driven customer insights extraction. 

While persona-based LLMs help businesses, their ability to understand persona and generate meaningful insights needs to be researched. Jiang et al. \cite{jiang2023personallm}, through a case study, investigate whether LLMs can generate content similar to assigned personas by simulating different personalities using the Big Five personality model. Their findings demonstrate that LLMs can adjust their output to match the behavior of assigned personas.

\subsection{Positioning of this Research in the Context of Related Work}
\label{section3.5}

The reviewed literature from Sections \ref{section3.1} to \ref{section3.4} highlights major advancements in persona creation and usage. This includes both traditional methods that do not use LLMs and newer methods that use LLMs. However, some limitations make it difficult to apply these methods in real-world businesses, such as the construction equipment manufacturing industry.

While studies such as \cite{barandoni2024automating} have compared prompting methods such as few-shot and CoT reasoning for tasks like customer needs extraction, they did not focus on structured persona generation. Moreover, studies on persona generation such as \cite{de2023improved, salminen2024deus,goel2023preparing} typically relied on a single prompting method without comparing multiple approaches. This brings a gap in understanding which prompting method works best when generating personas from qualitative data, such as customer success stories. Existing studies on using personas mainly focus on general use cases, such as how LLMs understand personas \cite{panda2024llms}, identifying stereotypes in LLM responses \cite{cheng2023marked}, and the use of personas to improve personalization \cite{ha2024clochat}. However, there is limited research on how customer personas can support businesses where the customer attributes, such as challenges and needs, vary widely. Finally, the reviewed studies do not explore the integration of customer personas with retrieval systems like RAG. The authors of \cite{praveen2024crafting} mention in their limitations and future directions that techniques like RAG could be beneficial for retrieving consumer data. However, the use of RAG for persona-based analysis and insight generation remains unexplored, especially in helping R\&D engineers and stakeholders to interact efficiently with persona data.
 
This research addresses the identified limitations in existing studies by systematically comparing and evaluating different prompting methods using specific metrics. The evaluation will help determine the most effective prompting technique for persona generation. This will be further discussed in the methodology section. Additionally, integrating a RAG-based system with personas allows R\&D engineers and stakeholders to interact with customer data more easily. This ensures that the findings of this research are not only theoretically grounded but also practically applicable in real-world business scenarios.

\section{Method And Implementation}
\label{sec:method}

\subsection{Research Method}
This research adopts the Design Science Research Methodology (DSRM) as the primary research framework \cite{peffers2007design}. DSRM is especially suitable for studies in computer science where the focus is on the development and evaluation of innovative artifacts to solve real-world problems. As described in \cite{johannesson2014method}, the DSRM framework includes five main activities: 
\begin{enumerate}
    \item Problem Explication
    \item Requirements Definition
    \item Design and Development 
    \item Demonstration
    \item Evaluation
\end{enumerate}

\begin{figure}[ht!]
    \centering
    \includegraphics[width=1\textwidth]{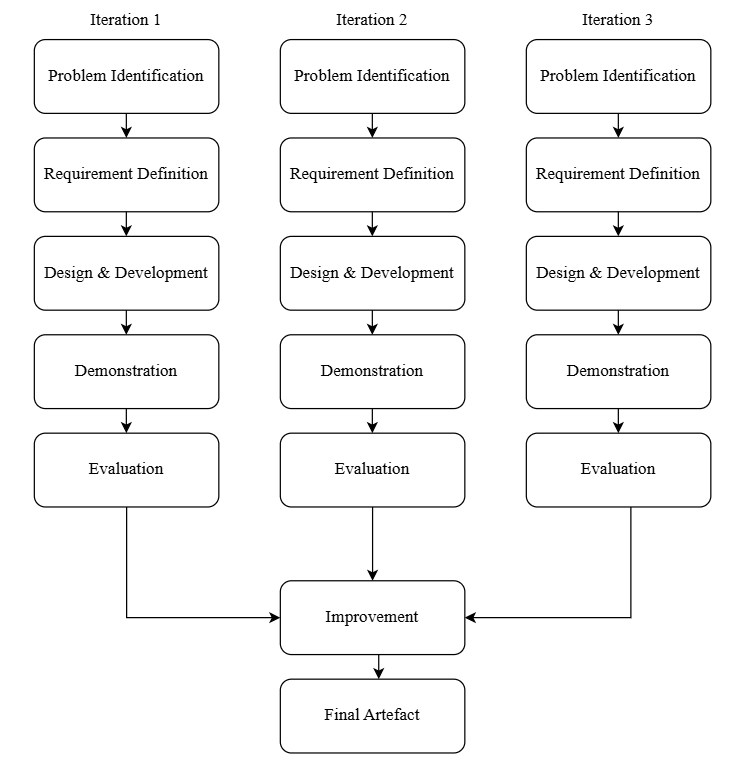}
    \caption{Research Method}
    \label{fig:researchmethod}
\end{figure}

To systematically address the research objectives, these activities were executed in three phases. Each iteration was built upon findings and evaluations from previous cycles, which improved the developed artifact. The artifact in this study is a conversational system designed to help stakeholders at VCE to query on customer persona. Figure~\ref{fig:researchmethod} depicts the research method followed in the study.

\textbf{Iteration 1: Initial Chatbot Development and Testing}
\begin{itemize}
\item \textbf{Problem Explication:} The initial problem identified was through a comprehensive literature review and discussions with stakeholders at VCE. The literature review explored existing research on customer personas, LLMs, RAG and prompting techniques. It helped providing a clear research gap regarding the integration of personas into RAG system and comparison of prompting method. The discussions with stakeholders highlighted the practical need for utilization of personas that can support various stakeholders to help make decisions faster.

\item \textbf{Requirements Definition:} Based on insights gained from the review of the literature and discussion with stakeholders, the key requirements were identified. This also included collecting and preparing relevant data and defining the chatbot's core functionalities.

\item \textbf{Design and Development:} The first artifact developed was a RAG-based chatbot that was integrated with verified customer personas provided by the VCE's Customer Experience team. Section~\ref{sec:rag_system}
discusses the process involved in building the RAG system.

\item \textbf{Demonstration:} The chatbot was deployed and demonstrated to a selected group of end-users. This enabled them to test and explore capabilities of the artifact.

\item \textbf{Evaluation:} An evaluation form was sent out to the end users who had the chance to explore the chatbot. The process involved in the evaluation is discussed in Section \ref{sec:initialevalchatbot} and the results of this section are discussed in Section \ref{sec:results_round1}. This feedback was then used as input for the second iteration.
\end{itemize}

\textbf{Iteration 2: Persona Generation and Comparison}
\begin{itemize}
    \item \textbf{Problem Explication:}
    Based on user feedback from Iteration 1, the problem identified was that the chatbot’s input data. The suggestion was to explore more data from customer success stories and more segment-specific information. Additional feedback included improved response accuracy and better handling of complex queries.
    
    \item \textbf{Requirements Definition:}
    As per the problem identified in the prior iteration, the requirement included gathering of additional personas, segment information and improvement of performance. In the literature study, there was also a gap in comparing the prompting technique for persona generation. This brings a requirement on studying how can different prompting techniques be used and evaluated.
    
    \item \textbf{Design and Development:}
    Personas were generated from customer success stories using two different prompting techniques. The development process involved in persona generation is elaborated in section \ref{sec:person_gen_process}.
    
    \item \textbf{Demonstration:}
    Personas created by both prompting techniques were clearly presented to evaluators alongside the original customer success stories. This helped them with a clear basis for comparison.
    
    \item \textbf{Evaluation:}
    A structured evaluation was conducted to statistically compare both prompting methods. The steps followed in the evaluation of personas are discussed in \ref{sec:persona_eval_process}. The statistical analysis tested which of the prompting generates better personas in terms of metrics (accuracy, relevance, and consistency). Based on these results, the generated synthetic persona was used to augment the knowledge base of the chatbot in the third iteration.
\end{itemize}

\textbf{Iteration 3: Improvement of the conversational system and the final evaluation}
\begin{itemize}
    \item \textbf{Problem Explication:} 
    Following the second iteration, the best-performing method of generating persona was determined. The other feedback from the first iteration included the addition of additional segment information.
    
    \item \textbf{Requirements Definition:} The requirement in this iteration was to improve the chatbot by updating the knowledge base. This was done by adding additional synthetic customer personas and additional segment-specific information. 
    
    \item \textbf{Design and Development:} The chatbot's knowledge base was expanded and refined by incorporating new data based on the initial feedback from iteration 1 and results from iteration 2.
    
    \item \textbf{Demonstration:} The improved chatbot was redeployed and demonstrated to various end-users for testing and exploration.

    \item \textbf{Evaluation:} A second evaluation round was conducted using a similar evaluation form from the iteration 1. This form assessed chatbot accuracy, usability, user satisfaction, and practical applicability. The purpose was to validate improvements from previous iterations and confirm the artifact effectively addressed the originally identified problem. The process involved in the evaluation is discussed in Section~\ref{sec:eval_augmented_chatbot} and the results of this section are discussed in Section \ref{sec:results_round2}.

\end{itemize}

\subsection{Overview of Data}
Three types of data were used in this study:

\begin{enumerate}
    \item \textbf{Customer Success Stories:} These are real-world narratives that illustrate how customers achieved positive outcomes using VCE's products or services. This data is publicly available on the VCE website \footnote{\url{https://www.volvoce.com/united-states/en-us/resources/customer-success-stories/}}. This contains text, images, and videos. The stories are categorized by product, application, or industry segment (e.g., agriculture, demolition, quarrying, and aggregates). In this study, only stories related to the quarrying and mining segments were used.

    \item \textbf{Verified Personas:} These personas were developed by the Customer Experience team at VCE through direct interviews with customers from different regions. This data is internal and confidential. It is accessible only to VCE employees. Table~\ref{tab:customer_attributes} contains the information included in each persona.

    \item \textbf{General information about the Quarry, Mining, and Aggregates segments:} This consists of textual data providing definitions, processes, and background information about the three industry segments. It is used as contextual knowledge to support understanding of the domain.

\begin{table}[h!]
\centering
\caption{Persona Attributes}
\label{tab:customer_attributes}
\begin{tabular}{p{4cm}p{8cm}}
\hline
\textbf{Attribute} & \textbf{Description} \\
\hline
Narrated Video & A video summarizing the persona's story \\
Name & The customer's name \\
Role & The job title or position \\
Number of Employees & Total employees in the customer's organization \\
Fleet Size & Size of the equipment fleet \\
Short Story & A brief background or narrative \\
What is Important & Key priorities or values of the customer \\
Challenges & Main issues faced by the customer \\
Expectations & What the customer expects from VCE \\
Buying Considerations & Factors that influence the customer's decisions \\
\hline
\end{tabular}
\end{table}

\end{enumerate}

\subsection{Data Preparation}
This section discusses the process involved in the data preparation. Each subsection describes the process involved for that specific data type.
 
 \subsubsection{Customer Success Story} The first type of data used was the Customer Success Stories, which served as input for generating synthetic customer personas. To extract these stories, web scraping was employed using the Python library Beautiful Soup\footnote{\url{https://pypi.org/project/beautifulsoup4/}}. Manual extraction of information from each webpage would have been time-consuming and error-prone. Hence, automated scraping was chosen to extract information. 
 
The process began by inspecting the HTML structure to identify the relevant tags that contained the main story content. Only the textual narrative was extracted. Non-relevant elements such as headings, image captions, videos, and figures were excluded.  A CSV file containing the URLs of selected (mining and quarrying segment) success stories was used as input for web scraping. From each URL, the script fetched the page content and extracted all paragraph (\textit{<p>}) elements within a specific section of the webpage (\textit{ div class "newsArticle-2023"}). The extracted text was then cleaned by removing extra spacing and manually adding the missing content that was not extracted during the scraping process.

\subsubsection{Verified Personas}
The second source of data is verified Personas provided by VCE. These personas are stored in an internal platform that is only accessible to VCE employees. Due to this restriction, web scraping was not a feasible option for this dataset. Thus, the persona data was manually copied from the internal website. All the textual content from each of the personas was extracted, excluding video material. Once the textual data was collected, it was converted into structured JSON files using a Python script. The data was then converted to JSON as it provides a structured, machine-readable format that enables integration with retrieval systems. Each JSON file represented a single persona and included key-value pairs corresponding to the persona attributes such as name, role, challenges, and expectations.

\subsubsection{General Information About the Quarry, Mining, and Aggregates Segments}
The third type of data used in this study was general textual information related to quarrying, mining, and aggregates. This content was used to provide contextual background for retrieval tasks. This will help the RAG system better understand industry-specific terminology and operations.

This data was completely textual content. To enhance readability and support LLM's, the whole text was manually split into small meaningful chunks based on meaningful topics. Each chunk was then converted into Markdown format to introduce structure and hierarchy within the documents. Headings, subheadings, and bullet points were added to clearly distinguish between concepts, definitions, and processes. Converting into markdown will help the system better recognize the relationships between different pieces of information.

\subsection{Generation Of Synthetic Customer Personas}
\label{sec:person_gen_process}

This section details the process of generating customer personas using customer success stories as input. In this study, personas were created using two different prompting techniques: few-shot prompting and CoT prompting. GPT-4o Mini was selected as the language model for persona generation. The model received both the prompt and the success story as input. The prompt designs were developed based on OpenAI's prompt engineering guidelines \cite{openai2025prompt}. Multiple iterations of prompts were refined and tested to improve the output quality. The refinement process involved experimenting with wording and adjusting the level of detail provided in the prompts.

\begin{figure}[h!]
    \centering
     \includegraphics[width=0.9\textwidth]
    {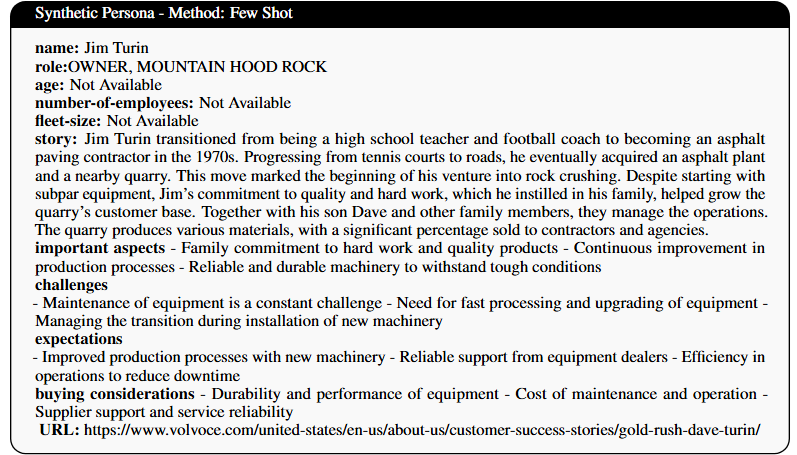}
    \caption{Example Synthetic Persona - Few Shot}
    \label{fig:persona-example-fewshot}
\end{figure}

In the few-shot prompting technique, the model was provided with three verified personas as examples. The complete prompt included system instructions, a task definition, an output structure format, and three example personas. The benefit of this method is that by using examples, the model will be able to recognize patterns and relationships between persona attributes. This helps the model generate structured and coherent outputs. Figure \ref{fig:persona-example-fewshot} is depiction of an example persona generated by few-shot prompting technique.

\begin{figure}[h!]
    \centering
     \includegraphics[width=0.9\textwidth]
    {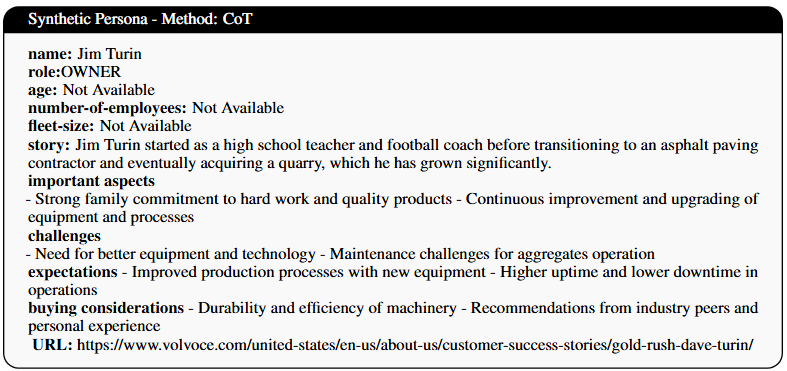}
    \caption{Example Synthetic Persona - Cot}
    \label{fig:persona-example-cot}
\end{figure}

The CoT prompting technique followed a different approach by guiding the model through an internal step-by-step reasoning process instead of directly generating persona attributes. The prompt included system instructions, a structured output format, and a reasoning process to improve information extraction. The model was instructed to first identify key details from the success story, then analyze the customer’s background and business context, extract challenges, expectations, buying considerations, and finally generate the structured persona. In this method, the model is encouraged to perform logical reasoning before output generation. Figure \ref{fig:persona-example-cot} is an illustration of a persona generated using this method.

\begin{figure}[h!]
    \centering
    \includegraphics[width=1\textwidth]{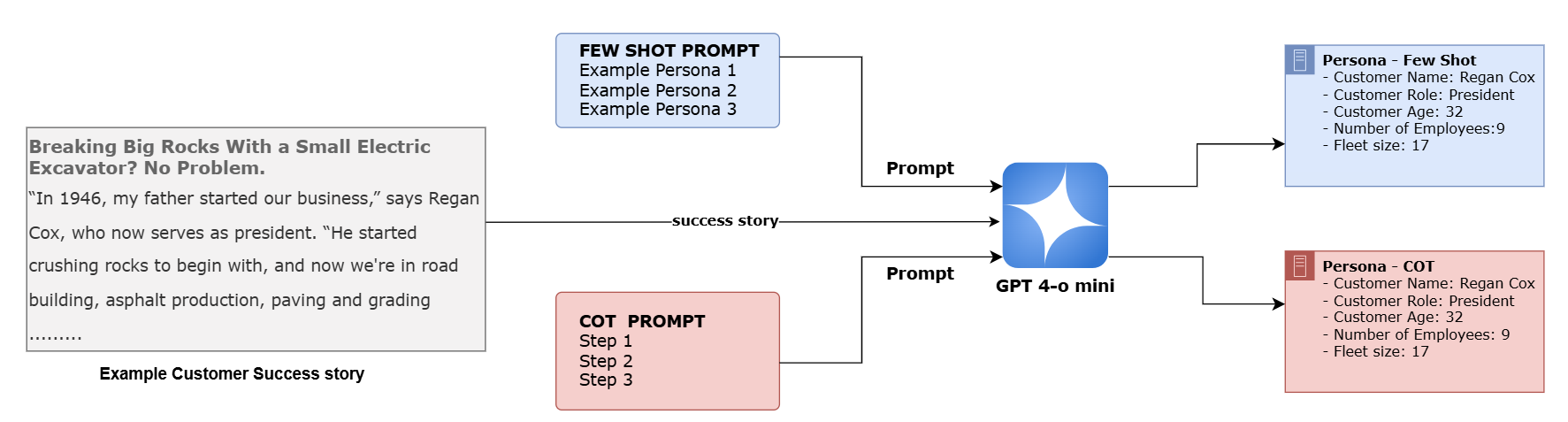}
    \caption{Persona Generation Process}
    \label{fig:personagen_process}
\end{figure}

For each of the personas generated by using the two prompting methods, the total time taken to generate personas in seconds and the total tokens consumed were also computed. The overall process involved in persona generation is illustrated in Figure~\ref{fig:personagen_process}  

\subsection{Building the RAG system}
\label{sec:rag_system}

The role of the RAG system is to act as a conversational agent that allows users to query information based on customer persona data and general information about different segments. The system consists of two main components: (1) \textbf{Retrieval Component} is responsible for storing, indexing, and retrieving relevant documents; and (2) \textbf{Generation Component} that is responsible for generating responses based on the retrieved content using LLM. The subsection below is a brief explanation of the design and implementation details of each component.

\subsubsection{Retrieval Component}

The retrieval component was built using Azure AI Search\footnote{\url{https://azure.microsoft.com/en-us/products/ai-services/ai-search}} that acts as a dedicated search engine and storage of data. The implementation process involved in building this component included the following steps:

\begin{enumerate}
    \item \textbf{Creating the Search Index:} The first step in constructing a retrieval system involves creating a search index. The index schema was designed to accommodate both structured persona data (*.JSON format) and unstructured general information (*.txt format). The schema contained the below fields:
    \begin{itemize}
        \item \texttt{id}- A unique identifier for each document.
        \item \texttt{title}- The name of the document.
        \item \texttt{category}- The type of document (e.g. "persona" or "general information" )
        \item \texttt{content}- The complete data in textual format that is to be searched and retrieved.
        \item \texttt{content\_vector}- A high dimensional vector representation of the document for similarity-based retrieval.
    \end{itemize}
    
While creating the search index, three search techniques were configured for efficient content retrievals:

\begin{itemize}
    \item \textbf{Keyword Search:} This type of search performs lexical matching based on the exact words in the query. It allows filtering and ranking documents using traditional search techniques.
    \item \textbf{Semantic Search:} This approach improves ranking by understanding the meaning of the query and prioritizing documents based on contextual relevance rather than just exact word matches. The \textit {content} field was set as the primary ranking factor to ensure meaningful results.
    \item \textbf{Vector Search:} This type of search was implemented using the Hierarchical Navigable Small World (HNSW) algorithm for Approximate Nearest Neighbor (ANN) retrieval. It enables searching for semantically similar documents using vector embeddings, even when the query and the document do not share exact words.
\end{itemize}

\item \textbf{Uploading Documents to the Index:} After the index was created, the next step involved uploading the documents to the index that was created. This process began by loading the data and extracting textual content from the data. The textual content was then converted into embeddings using an embedding model named text-embedding-ada-002. These embeddings, along with the raw text, were then batch-uploaded into the index.

\end{enumerate}

\subsubsection{Generation Component}
The Generation Component is responsible for utilizing the content retrieved by the retrieval component to generate output for the user. This component was developed using GPT-4o Mini in combination with a hybrid search approach. Below are the implementation details of the processes involved in this component.

\begin{enumerate}
    \item \textbf{Integrating search index with hybrid search strategy:} To improve the quality of responses, a hybrid search approach was employed. Hybrid search combines the capabilities of both keyword-based search and vector-based search techniques. According to experiments by Microsoft \cite{azure2024hybrid}, hybrid search outperforms standalone keyword or vector-based search methods in retrieving relevant documents for question-answering systems. Due to this reason, the hybrid search was opted for this study.
    
    When a user submits a query, the query is first converted into an embedding using the embedding model. The hybrid search method is then applied to retrieve the top three most relevant documents from the search index. These documents are used as contextual input for the language model to perform the generation of an appropriate response.

    \item \textbf{System Message and Prompt Engineering:}
To ensure consistency, accuracy, and contextual relevance in the generated responses, a Prompty file \footnote{\url{https://prompty.ai/}} was created. This file contains a system message that defines specific role instructions, the tone for responses, and detailed guidelines for answering the questions.

    \item \textbf{Final Response Generation:}
After retrieving the relevant documents, the GPT-4o Mini model synthesizes the final response by integrating the retrieved documents, system message, and conversation history to ensure coherent and context-relevant output.

\end{enumerate}

\begin{figure}[h!]
    \centering
    \includegraphics[width=1\textwidth]{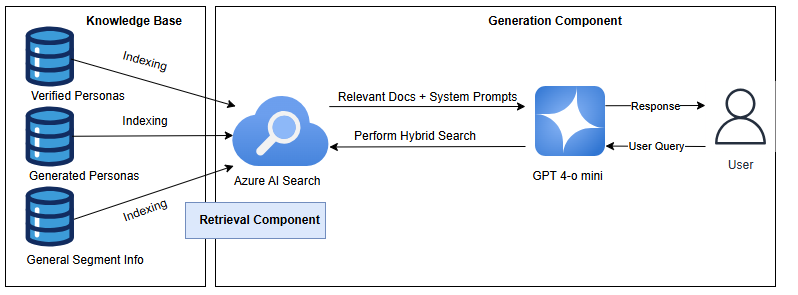}
    \label{fig:chatbot_process}
\end{figure}

Figure \ref{fig:chatbot_process} shows the overall process of the chatbot system, highlighting the flow from data indexing and retrieval to response generation using a hybrid search strategy.

\subsection{Initial Evaluation of the Conversational System}
\label{sec:initialevalchatbot}

This section describes the process involved in the initial user evaluation to assess the effectiveness of the persona-based chatbot system. The goal of this evaluation is to understand how the chatbot supports decision-making and contributes to automating customer-facing processes. For this stage of evaluation, the developed chatbot was integrated with verified customer personas and segment-specific information. The system was then deployed using Azure Web App for user interaction \footnote{\url{https://azure.microsoft.com/en-us/products/app-service/web}}. The participants included people from relevant business functions, such as R\&D, marketing, and customer representatives. After interacting with the system, they were asked to provide feedback through an evaluation form.

\subsubsection{Evaluation Design}

To systematically assess the ability of the chatbot to support decision-making and process automation, five questions were opted. The type of questions includes multiple choice, Likert's scale, and 1-10 rating scale. Table \ref{tab:chatboteval1} presents the questions in the evaluation form along with their purposes and types of responses.

\begin{table}[h!]
\centering
\caption{Initial Evaluation Questions, Purpose and Type}
\label{tab:chatboteval1}
\begin{tabular}{|p{5.5cm}|p{5.5cm}|p{2.5cm}|}
\hline
\textbf{Question} & \textbf{Purpose} & \textbf{Response Type} \\
\hline
How would you rate the chatbot's ability to provide accurate answers? & Evaluate the overall accuracy of the chatbot in providing  relevant answers& 1-10 scale \\
\hline
Does the chatbot correctly interpret and respond to complex queries (e.g., providing details on customer personas)?
& Measures the chatbot's capability to handle complex queries.& Likert Scale\\
\hline
Does the chatbot provide clear and concise answers?
& Assesses clarity in responses.& Likert Scale\\
\hline
How well does the chatbot align with your business needs?
& Evaluates the chatbot’s relevance and usefulness in business contexts.& Likert Scale\\
\hline
Do you believe the chatbot has reduced the workload for human support teams?

& Understand the impact of chatbot in automation and improving efficiency.& Multiple Choice\\
\hline
\end{tabular}
\end{table}

In addition to these questions, participants were also asked to provide open-ended comments to elaborate on their experience, share specific feedback, or suggest improvements. This combination of quantitative and qualitative feedback designed helped in providing insights into how well the system aligns with user expectations, business needs, and opportunities for automation. Data from these responses was analyzed using descriptive statistical analysis and qualitative thematic analysis. The findings are presented in section \ref{sec:results_round1} in the chapter \ref{sec:results}.

\subsection{Evaluation of Generated Personas} 
\label{sec:persona_eval_process}
This section presents the methodology used to evaluate the two types of prompting techniques. The aim of this evaluation is to identify the optimal prompting technique to produce personas. 

\subsubsection{Evaluation Design}
A total of 24 customer success stories were used to generate personas. To reduce the time and effort for evaluators, a random subset of five stories was selected for the evaluation. Each evaluator read the full customer success story before reviewing two anonymized personas. The order in which the personas were presented was randomized to minimize the bias. A Microsoft Form \footnote{\url{https://forms.office.com/}} was used to collect binary feedback (Yes/No) on each of the evaluation metrics.

\subsubsection{Metrics Used}
The metrics used in this study were adapted from literature \cite{salminen2024deus, sun2024persona, goel2023preparing} that evaluated personas. Each of the metrics was evaluated using binary response. The choice of binary metrics was to reduce ambiguity and speed up the evaluation process.

Table \ref{tab:metrics} presents the questionnaire used for evaluating each metric, along with a brief description of what each metric assesses.

\begin{table}[h!]
\centering
\caption{Metrics and Definition}
\label{tab:metrics}
\begin{tabular}{|p{2cm}|p{6cm}|p{6cm}|}
\hline
\textbf{Metric Name} & \textbf{Questionnaire} & \textbf{Description} \\
\hline
Completeness & Does the persona include all the important details (like role, challenges, expectations etc.) from the customer success story to fully understand the customer? & Evaluates whether the persona captures all key customer insights needed for understanding. \\
\hline
Relevance & Does the persona focus only on the relevant and important details from the customer success story? & Assess whether the persona includes only important details from the source story, avoiding any irrelevant or redundant information. \\
\hline
Consistency & Does the persona add any incorrect or made-up information that is not in the customer success story? & Checks if the persona introduced incorrect, fabricated, or contradictory information.\\
\hline
\end{tabular}
\end{table}

\subsubsection{Participants}
The evaluation was conducted with professionals from VCE who are familiar with customers, products, and services. The evaluators included Customer Solution Strategists, Research Engineers, and Project Managers. 

\subsubsection{Analysis Method}

A formal hypothesis testing approach was adopted to determine whether differences between prompting methods were statistically significant. For each prompting method and for each metric, the following hypotheses were defined:
\begin{itemize}
    \item \textbf{Null Hypothesis ($H_0$)} : There is no significant difference between the two prompting methods in terms of the metrics used for evaluation.
    \item \textbf{Alternative Hypothesis ($H_1$)}: There is a significant difference between the two prompting methods in terms of the metrics used for evaluation.
\end{itemize}

The McNemar test \cite{pembury2020effective} was selected because it is specifically designed for paired nominal (categorical) data. It is commonly used when the same subjects are exposed to two conditions, and their binary responses (e.g., Yes/No) are analyzed for shifts between the two conditions \cite{pembury2020effective}. In this research, it is used to determine if one prompting method significantly outperformed another across the three binary evaluation metrics. This test uses a 2$\times$2 contingency table based on paired binary responses for each persona pair. Only the discordant pairs, where evaluators responded differently for the two methods  are used to compute the test statistic. A separate contingency table is constructed for each evaluation metric. Table \ref{tab:contigtableex} presents an example contingency table.

\begin{table}[h!]
\centering
\caption{Example Contingency Table}
\label{tab:contigtableex}
\begin{tabular}{|p{4cm}|p{4cm}|p{4cm}|}
\hline
\textbf{} & \textbf{Method: CoT - Yes} & \textbf{Method: CoT - No} \\
\hline
\textbf{Method: Few-Shot - Yes} & a & b \\
\hline
\textbf{Method: Few-Shot - No} & c & d \\
\hline
\end{tabular}
\vspace{0.5em}

$a$ = Both methods are rated as 'Yes' \\
$b$ = Method Few-Shot is rated as 'No' and Method CoT is rated as 'Yes' \\
$c$ = Method Few-Shot is rated as 'Yes' and Method CoT is rated as 'No' \\
$d$ = Both methods are rated as 'No'
\end{table}
The McNemar test statistic is defined as \( \chi^2 = \frac{(b - c)^2}{b + c} \). The test statistic value which we obtain follows a chi-square distribution with one degree of freedom. From this value, a p-value is calculated and used to assess whether the observed difference is statistically significant. A p-value below 0.05 indicates a significant difference between the two prompting methods for the given evaluation metric.

\subsection{Evaluation of Augmented Chatbot with Synthetic Personas}
\label{sec:eval_augmented_chatbot}

This section discusses the methodology used to evaluate the impact of augmenting the chatbot’s knowledge base with synthetic personas and additional segment-specific information. The objective was to assess whether these enhancements improved the chatbot's performance in terms of accuracy, usability, and decision-making capabilities.

\textbf{System Updates and Evaluation Design}

Based on the insights from Section \ref{sec:initialevalchatbot} (Initial Evaluation) and Section \ref{sec:persona_eval_process} (Persona Generation Evaluation), the conversational system was updated to improve its overall performance. "Feedback from the initial evaluation highlighted the need for additional segment data and more personas derived from customer success stories. As a result, the knowledge base was expanded with newly generated synthetic personas and additional segment-specific information.

Based on the findings in Section \ref{sec:persona_eval_process}, the best-performing prompting technique for persona generation was selected. The personas generated by this method were then added to the Azure AI Search index, replacing the initial dataset. Plus, the system prompt that was used to guide the chatbot’s responses was also revised. This was to improve the accuracy, particularly for complex or context-rich queries.

Once these changes were implemented, the system was tested and redeployed. The updated version of the chatbot was made available for user testing. To ensure consistency and for direct performance comparison, the same evaluation method, participant group, and questionnaire from the initial evaluation were used again.
The collected responses were analyzed using the same methods as in the initial evaluation. Descriptive statistical analysis was used to compare quantitative results. This approach enabled direct comparison between the initial chatbot (with verified personas) and the updated version (with synthetic personas). The findings from this evaluation are presented in Chapter \ref{sec:results_round2}.

\section{Results}
\label{sec:results}

This section presents the results of this study, including the evaluation of generated personas and the performance of the persona-based chatbot. The results are organized based on the research questions.

\subsection{Results for Research Question 1: Effectiveness of the Persona-Based Chatbot}
\label{sec:results_round1}

This subsection presents the findings related to the evaluation of the persona-based chatbot conducted with eight stakeholders from relevant business functions such as R\&D, marketing, and customer relations. The evaluation was focused on five aspects, such as accuracy of the answers, the ability to handle complex queries, clarity of the responses, alignment with business needs, and impact on workload reduction.

\subsubsection{Quantitative Results}
\begin{figure}[h!]
    \centering
     \includegraphics[width=0.8\textwidth]
    {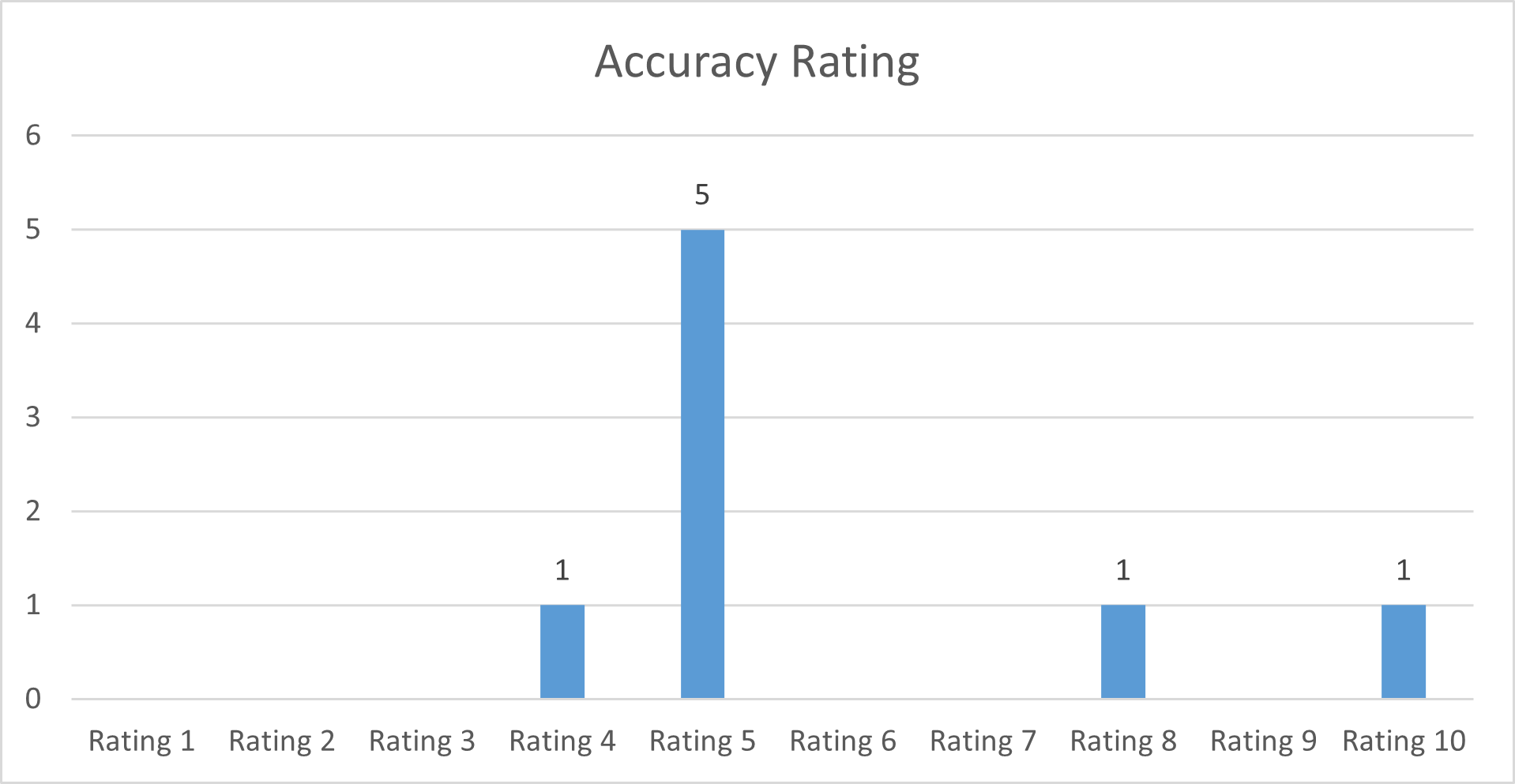}
    \caption{Distribution of the accuracy rating}
    \label{fig:accuracyround1}
\end{figure}

Participants were asked to provide an overall rating for the ability of the chatbot to provide accurate answers on a scale from 1 (Very Poor) to 10 (Excellent). The average rating across all evaluators was 5.88. The range of ratings from all evaluators was 4 to 10, with the majority giving a rating of 5. Figure~\ref{fig:accuracyround1} is a bar chart illustrating the distribution of accuracy ratings.

\begin{figure}[h!]
    \centering
     \includegraphics[width=0.8\textwidth]
    {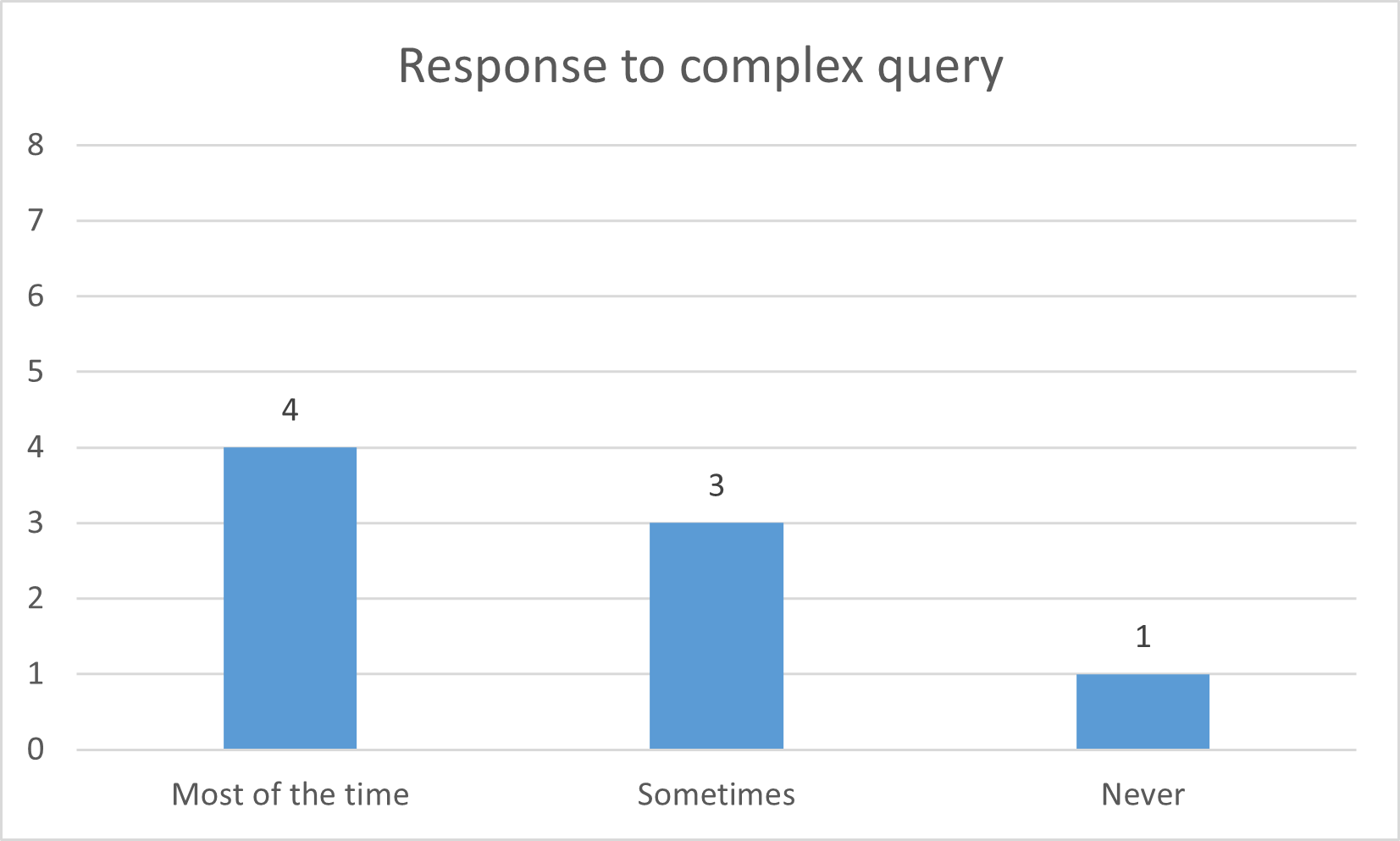}
    \caption{Ability of the system to provide response to complex query}
    \label{fig:complexqueryround1}
\end{figure}

The ability of the chatbot to interpret and respond to complex queries such as providing details on customer persona was measured using a Likert scale. The majority of the participants indicated that the chatbot responded correctly "most of the time", while three participants selected "sometimes", and one participant reported "never". Figure~\ref{fig:complexqueryround1} presents a bar chart depicting these findings.

\begin{figure}[h!]
    \centering
     \includegraphics[width=0.8\textwidth]
    {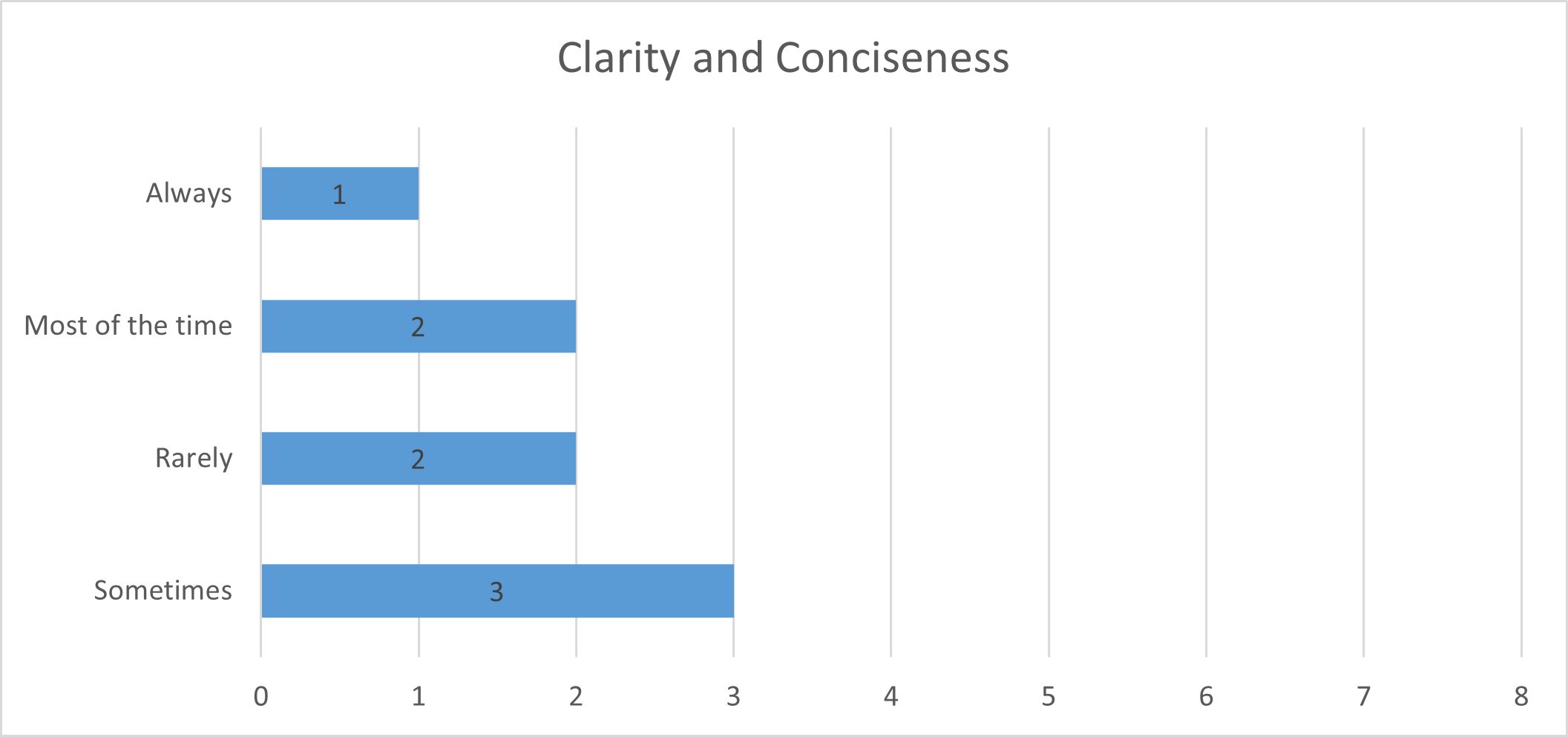}
    \caption{Ability of the system to provide clear and concise response}
    \label{fig:clearanswer_round1}
\end{figure}

Regarding the ability of the conversational system to provide clear and concise responses, 3 users reported that the system provides clear and concise results only sometimes. This indicates some inconsistency in the clarity and conciseness of the responses. Figure \ref{fig:clearanswer_round1} shows the distribution of the responses in a bar chart.

\begin{figure}[h!]
    \centering
     \includegraphics[width=0.8\textwidth]
    {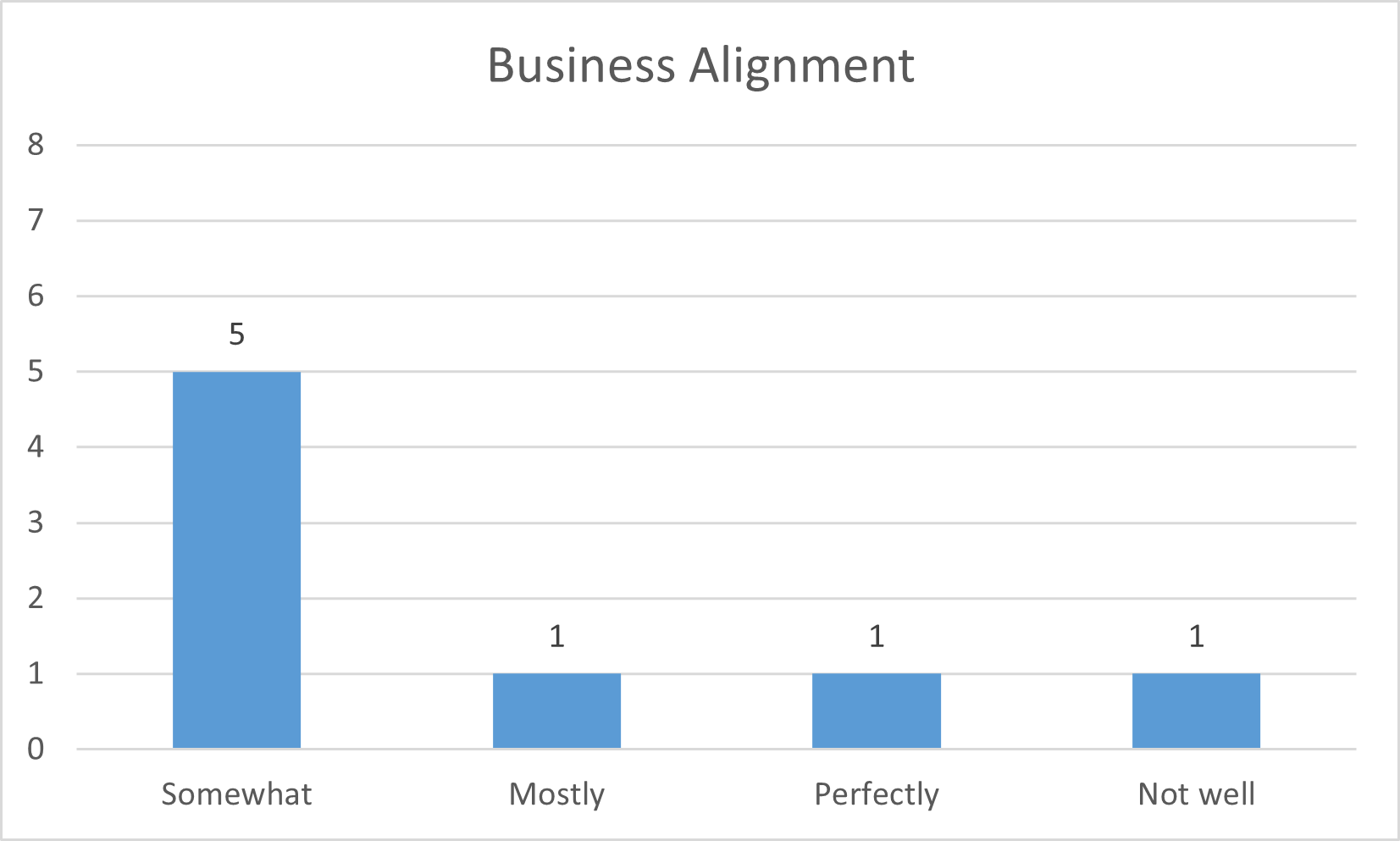}
    \caption{Alignment of system in business need}
    \label{fig:businessalignment_round1}
\end{figure}

The impact of the system in business contexts can be seen largely positive. Only one evaluator rated the system as not useful. The remaining participants responded positively. 62.5\% users rated it as "somewhat needed", suggesting that while the chatbot addressed business needs to some extent, further alignment and improvements are required. Figure~\ref{fig:businessalignment_round1} is a bar chart showing the distribution of the various ratings regarding the alignment of business needs.

\begin{figure}[h!]
    \centering
     \includegraphics[width=0.8\textwidth]
    {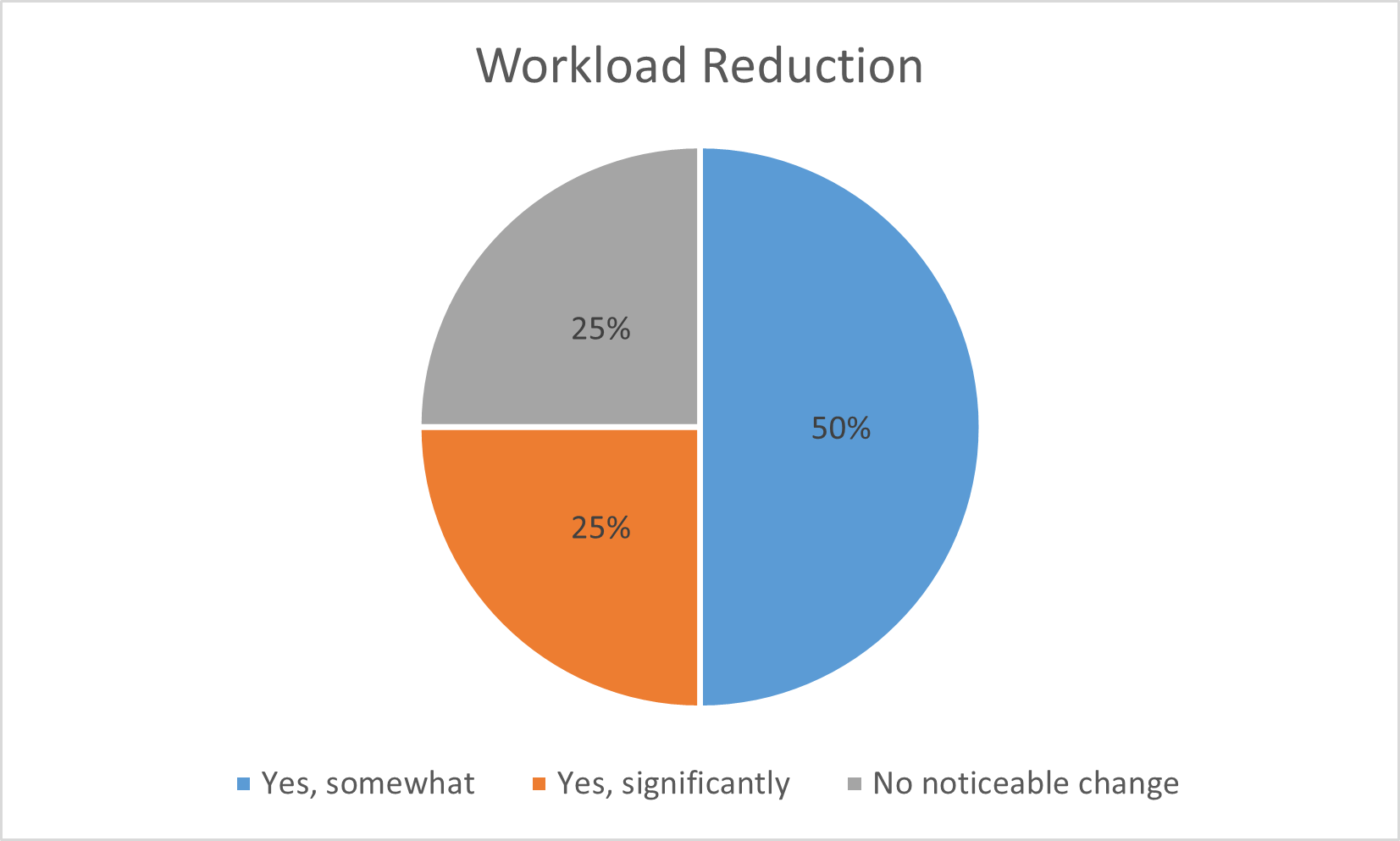}
    \caption{Workload reduction and improvement in automation}
    \label{fig:workloadreduction_round1}
\end{figure}

When evaluating the potential of the system to reduce the workforce, 75\% of the participants believe that it will reduce their workload and improve automation. Figure~\ref{fig:workloadreduction_round1} is a pie chart depicting this distribution.

\subsubsection{Qualitative Results}
In addition to the five evaluation questions, participants were asked to provide open-ended feedback on their experience with the persona-based chatbot. Three main key themes were identified from their responses. 

Primarily, the participants emphasized the importance of improving the quality of the data used. They suggested that integrating customer success stories and incorporating a more diverse range of customer data  could significantly enhance the chatbot’s utility. Another observation was that the responses sometimes felt too generalized and lacked specific insights, making them indistinguishable from publicly available information. This highlights that the system needs to provide deeper, more personalized answers based on additional customer datasets. Secondly, participants pointed out the necessity of improving segment-specific information within the chatbot’s knowledge base. Finally, the participants also identified several technical areas for improvement. These included better handling of complex queries, improved accuracy in responses, faster response times, and a more natural, human-like conversational style. Participants also suggested better integration with internal systems and the inclusion of sources for the information provided in responses.

\subsubsection{Summary of Findings}
The initial round evaluation of the persona-based chatbot demonstrated a moderate overall effectiveness in supporting decision-making and automation within the construction industry. From the quantitative results, the system received an average overall accuracy of 5.88 out of 10. While it generally aligned with business needs for  most of the users, there remains significant room for improvement. From the feedback on the ability of the system to reduce workload and automation, it can be concluded that the persona chatbot has contributed to reducing human workload. Along with these quantitative findings, the qualitative feedback highlighted the need for broader and more diverse data integration from customer success stories, enhanced segment-specific information, improved handling of complex queries, and more human-like interactions. Overall, these findings provided critical insights for guiding the refinement and improvement of the chatbot in the next iterations.

\subsection{Results for Research Question 2: Persona Generation and Prompting Techniques}
\label{sec:results_prompting}

This section presents the results obtained from comparing synthetic personas generated by two prompting techniques. 

\subsubsection{Quantitative Results}
 In this research, the synthetic persona was generated using the customer success story as input. The process involved in the generation of the persona is described in the section \ref{sec:person_gen_process}. The evaluation was carried out by three expert evaluators (n=3), each of whom reviewed a total of 5 personas. Each anonymized persona was assessed using three binary metrics: completeness, relevance, and consistency. The responses were analyzed using the McNemar's test to identify if there are statistically significant differences between the two prompting methods. In addition, efficiency metrics such as average generation time (seconds) and token usage were recorded.

\begin{figure}[h!]
    \centering
     \includegraphics[width=0.8\textwidth]
    {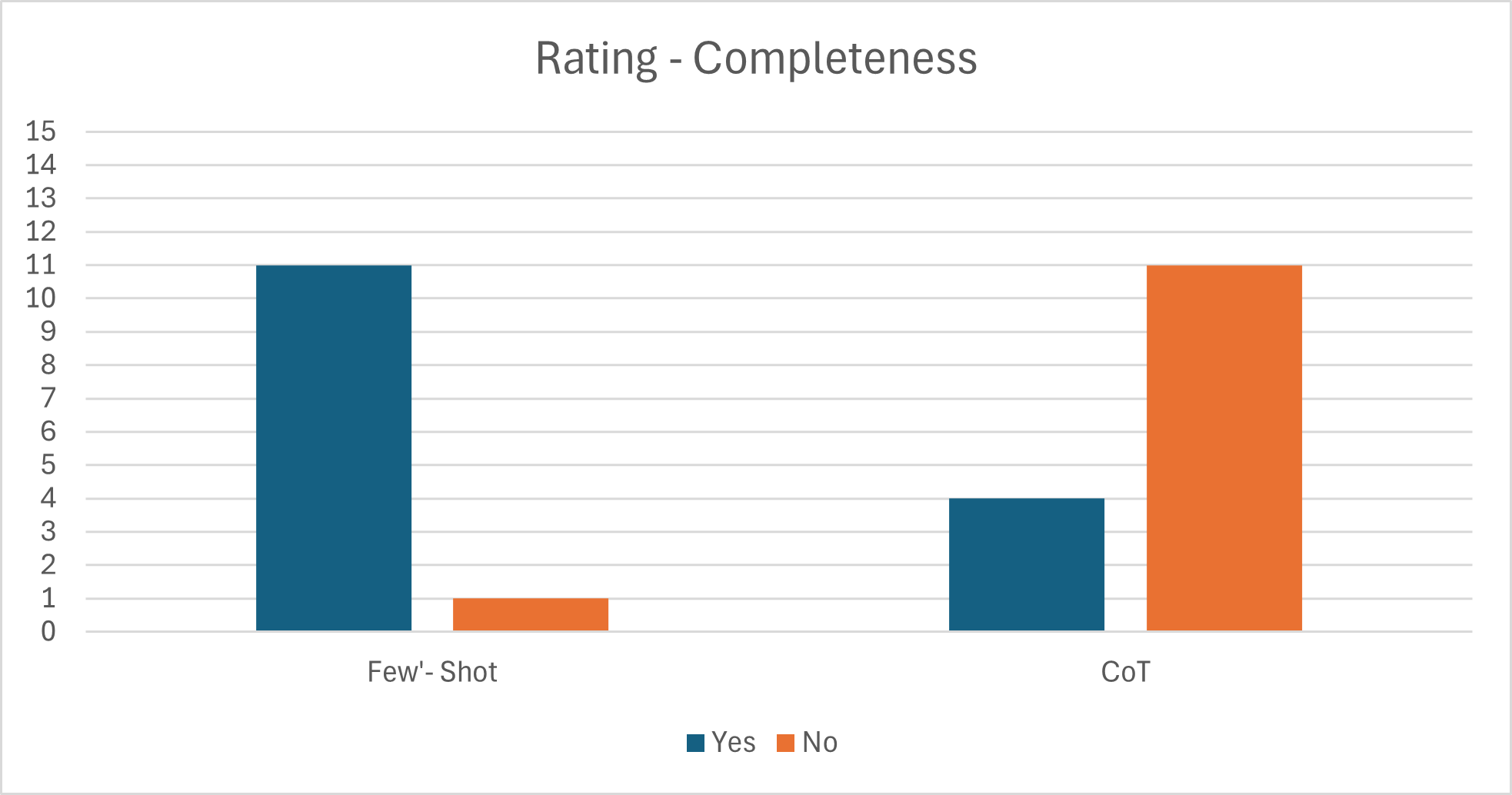}
    \caption{Comparison of the metrics- Completeness}
    \label{fig:completeness_graph}
\end{figure}

\textbf{Completeness:} The completeness metric evaluated whether the persona captured all important details (e.g., role, challenges, expectations) from the customer success story. The McNemar test produced a test statistic of 1.0 and a p-value of 0.0063, indicating a statistically significant difference between the two prompting methods. As shown in contingency table \ref{tab:contigtablecompleteness}, in 11 cases evaluators rated the Few-Shot persona as complete and the CoT persona as not complete, and in only 1 case the opposite occurred. This result clearly indicates that Few-Shot prompting outperformed CoT prompting in terms of completeness. Figure \ref{fig:completeness_graph} presents a bar chart that illustrates the distribution of the evaluator ratings for this metric.

\begin{table}[h!]
\centering
\caption{Contingency Table for Completeness Metric}
\label{tab:contigtablecompleteness}
\begin{tabular}{|c|c|c|}
\hline
\textbf{} & \textbf{CoT: Yes} & \textbf{CoT: No} \\
\hline
\textbf{Few-Shot: Yes} & a = 3 & b = 11 \\
\hline
\textbf{Few-Shot: No}  & c = 1 & d = 0 \\
\hline
\end{tabular}
\end{table}

\textbf{Relevance:} Relevance was assessed to determine whether the persona focused only on the important and relevant details from the source material. The test yielded a p-value of 0.6250, suggesting that there is no significant difference between the two prompting techniques. Table \ref{tab:contigtablerelevance} is the contingency table for the relevance metrics. While there were some variances in individual ratings, the differences were not statistically significant.

\begin{table}[h!]
\centering
\caption{Contingency Table for Relevance Metric}
\label{tab:contigtablerelevance}
\begin{tabular}{|c|c|c|}
\hline
\textbf{} & \textbf{CoT: Yes} & \textbf{CoT: No} \\
\hline
\textbf{Few-Shot: Yes} & a = 11 & b = 3 \\
\hline
\textbf{Few-Shot: No}  & c = 1 & d = 0 \\
\hline
\end{tabular}
\end{table} 

\textbf{Consistency:} The consistency metric evaluated whether the persona introduced incorrect or fabricated information. The test result for this metric was a p-value of 0.2500, indicating no significant difference between the two prompting approaches. Table~\ref{tab:contigtableconsistency} is the contingency table for this metric.

\begin{table}[ht]
\centering
\caption{Contingency Table for Consistency Metric}
\label{tab:contigtableconsistency}
\begin{tabular}{|c|c|c|}
\hline
\textbf{} & \textbf{CoT: Yes} & \textbf{CoT: No} \\
\hline
\textbf{Few-Shot: Yes} & a = 1 & b = 3 \\
\hline
\textbf{Few-Shot: No}  & c = 0 & d = 11 \\
\hline
\end{tabular}
\end{table}

\begin{figure}[h!]
    \centering
     \includegraphics[width=0.8\textwidth]{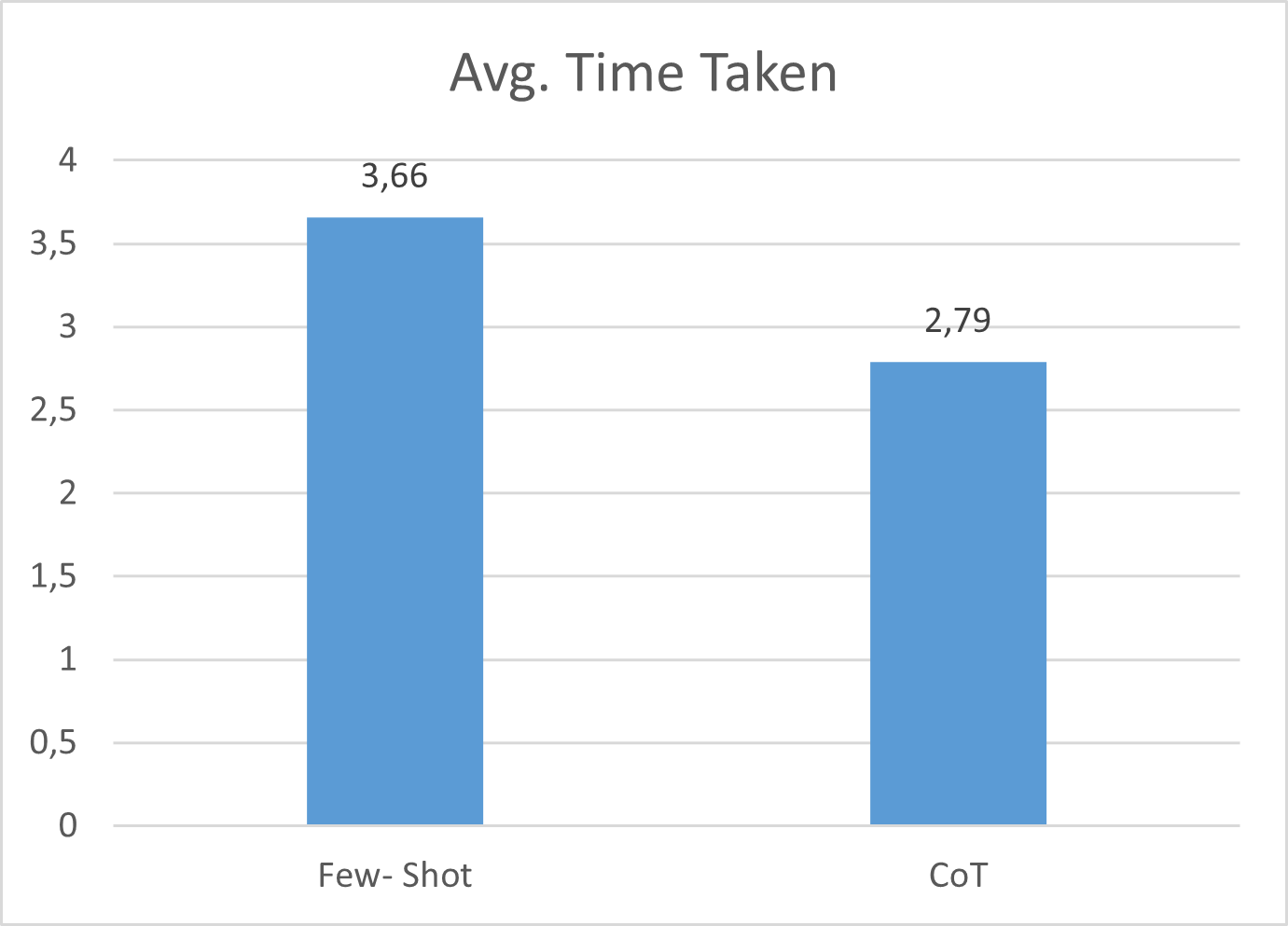}
    \caption{Average Time Taken}
    \label{fig:avgtime}
\end{figure}

\begin{figure}[h!]
    \centering
     \includegraphics[width=0.8\textwidth]{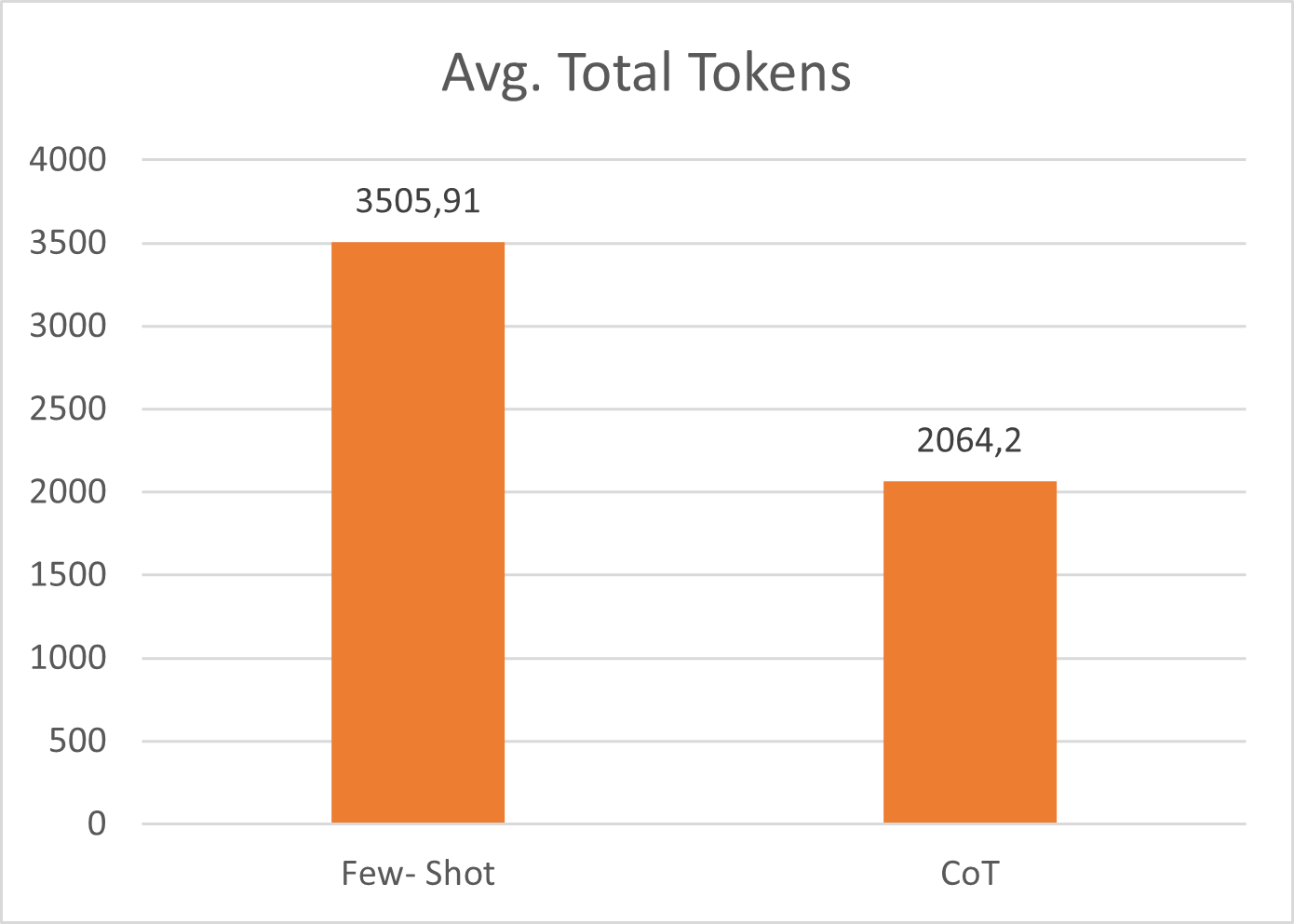}
    \caption{Average Tokens Consumed}
    \label{fig:avgtokens}
\end{figure}

\textbf{Efficiency Metrics:} In addition to qualitative performance, the two prompting methods were evaluated based on average generation time and token usage. Few-Shot prompting had an average generation time of 3.66 seconds and used 3505.91 tokens. While for CoT prompting,  the average generation time was 2.79 seconds and the total average tokens consumed was 2064.2. Figure~\ref{fig:avgtime} and Figure \ref{fig:avgtokens} show the average time and average token usage across all personas. These results indicate that CoT prompting outperformed Few-Shot as it required less time and fewer tokens, making it superior and computationally more efficient.

\subsubsection{Summary of Findings}

The evaluation revealed that Few-Shot prompting significantly outperformed CoT prompting in terms of completeness. Evaluators generally rated Few-Shot personas as more complete than those generated using CoT. For the other two quality metrics, relevance and consistency, no statistically significant differences were found between the prompting methods. However, when evaluated from an efficiency perspective, CoT prompting proved to be both faster and more resource-efficient. The personas generated using the CoT method use fewer tokens and have shorter response times compared to those generated with Few-Shot prompting. Given its superiority in efficiency, CoT prompting was selected for the second iteration of the system. Table \ref{tab:summary_prompting_eval} summarizes the comparative performance of the two prompting methods in terms of quality and efficiency.

\begin{table}[h]
\centering
\caption{Summary of Evaluation of Prompting Techniques}
\label{tab:summary_prompting_eval}
\resizebox{\textwidth}{!}{%
\begin{tabular}{|c|c|c|c|c|c|}
\hline
\textbf{Prompting Method} & \textbf{Completeness} & \textbf{Relevance} & \textbf{Consistency} & \textbf{Avg Time (s)} & \textbf{Avg Total Tokens} \\
\hline
Few-Shot & \multirow{2}{*}{Statistically significant} & \multirow{2}{*}{Statistically insignificant} & \multirow{2}{*}{Statistically insignificant} & 3.66 & 3505.91\\
\cline{1-1} \cline{5-6}
CoT      &                                            &                               &                                & 2.79 & 2064.2 \\
\hline
\end{tabular}%
}
\end{table}

\subsection{Results for Research Question 3: Impact of Knowledge Base Augmentation}
\label{sec:results_round2}

This section presents the results after augmenting the chatbot's knowledge base with synthetic personas and segment-specific information. 12 stakeholders from business functions such as R\&D, marketing, and customer relations had participated in the evaluation. The evaluation was focused on three primary aspects: accuracy of the responses, ability to handle diverse queries, and overall usefulness of the chatbot in business needs.

\subsubsection{Quantitative Results}
\begin{figure}[h!]
    \centering
     \includegraphics[width=0.8\textwidth]
    {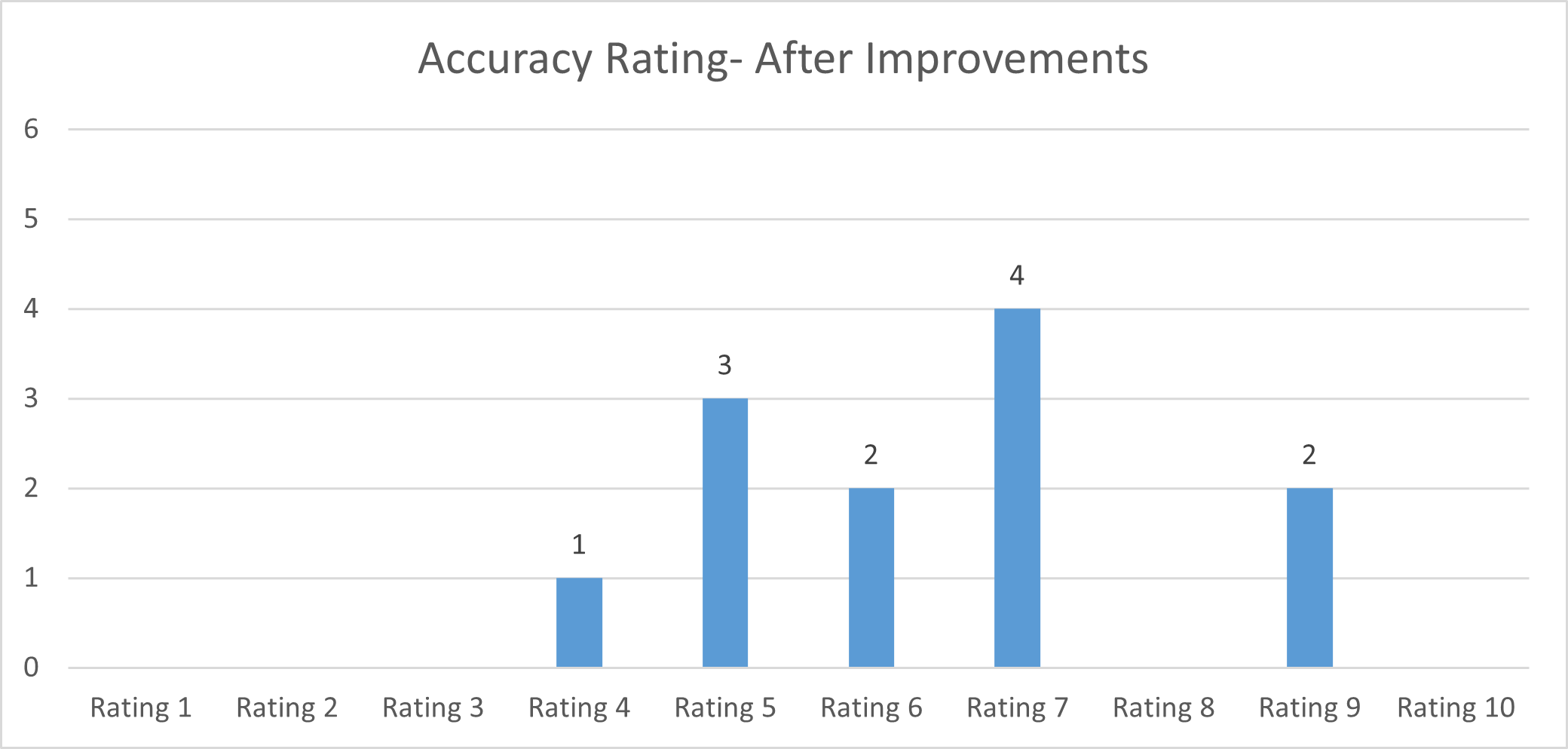}
    \caption{Distribution of the accuracy rating -  post improvements}
    \label{fig:accuracyround2}
\end{figure}

Participants were asked to assess the impact of the augmented knowledge base on the accuracy of responses, using a scale from 1 (Very Poor) to 10 (Excellent). The average rating across all evaluators was 6.42. This shows a slight improvement from the previous round of chatbot evaluation where the average rating was 5.88. The range of ratings was from 4 to 8. Most of the evaluators gave ratings of 6 or above. This indicates more consistent performance. Figure~\ref{fig:accuracyround2} is a bar chart depicting the distribution of the accuracy rating post improvement.

\begin{figure}[h!]
    \centering
     \includegraphics[width=0.8\textwidth]
    {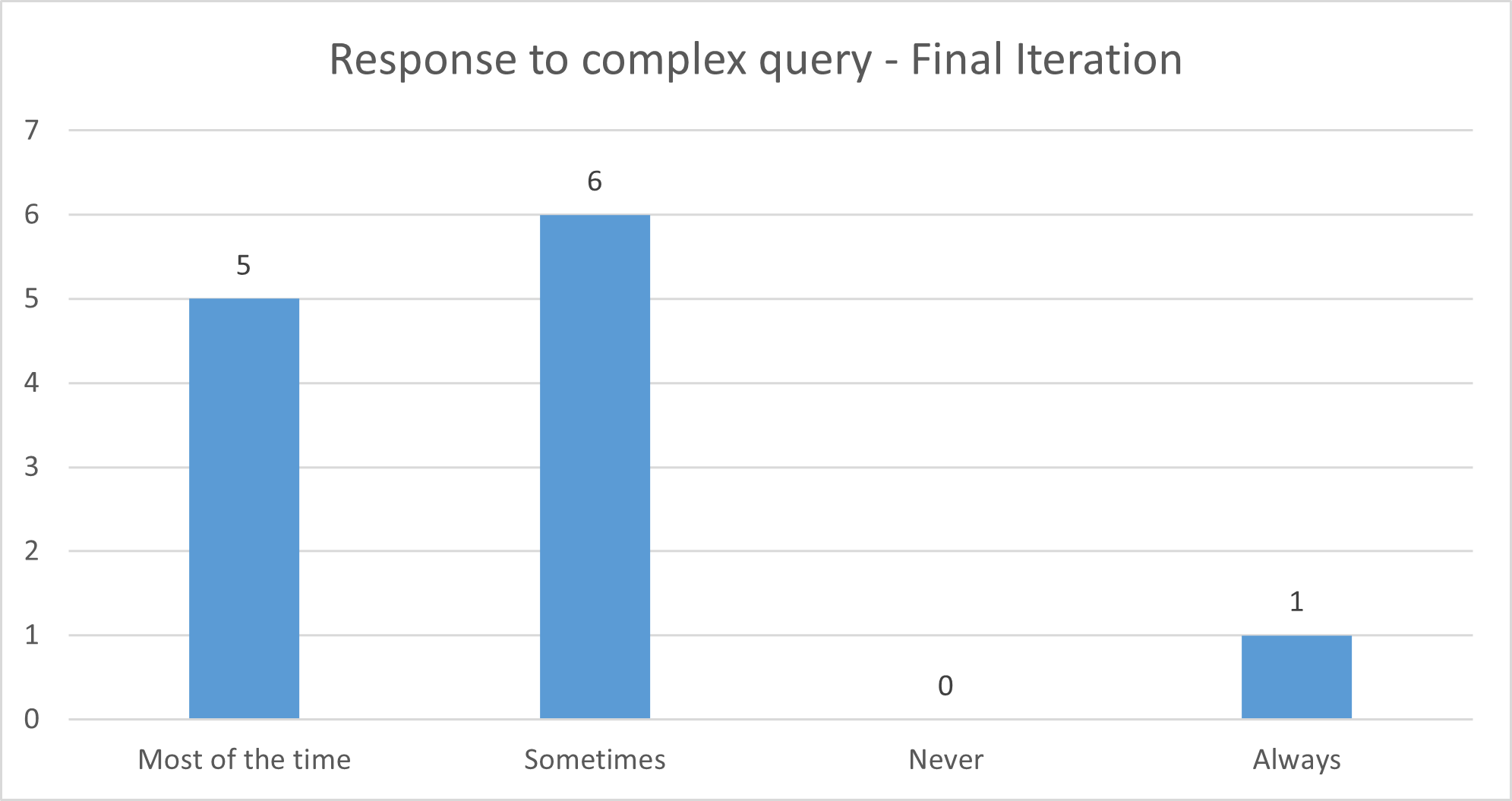}
    \caption{Ability of the system to provide response to complex query - post augmentation}
    \label{fig:complexqueryround2}
\end{figure}

To evaluate if the system was able to handle diverse and complex queries, participants were asked to assess the system's ability to respond to complex persona-related queries after data augmentation. 6 participants selected "sometimes", 5 of them selected "most of the time", and 1 selected "always". While the system showed balanced performance in handling complex queries, the responses like "sometimes" indicate some inconsistencies in providing comprehensive and contextually relevant answers. 
Figure~\ref{fig:complexqueryround2} is a bar graph that illustrates the response to handling complex queries.

\begin{figure}[h!]
    \centering
     \includegraphics[width=0.8\textwidth]
    {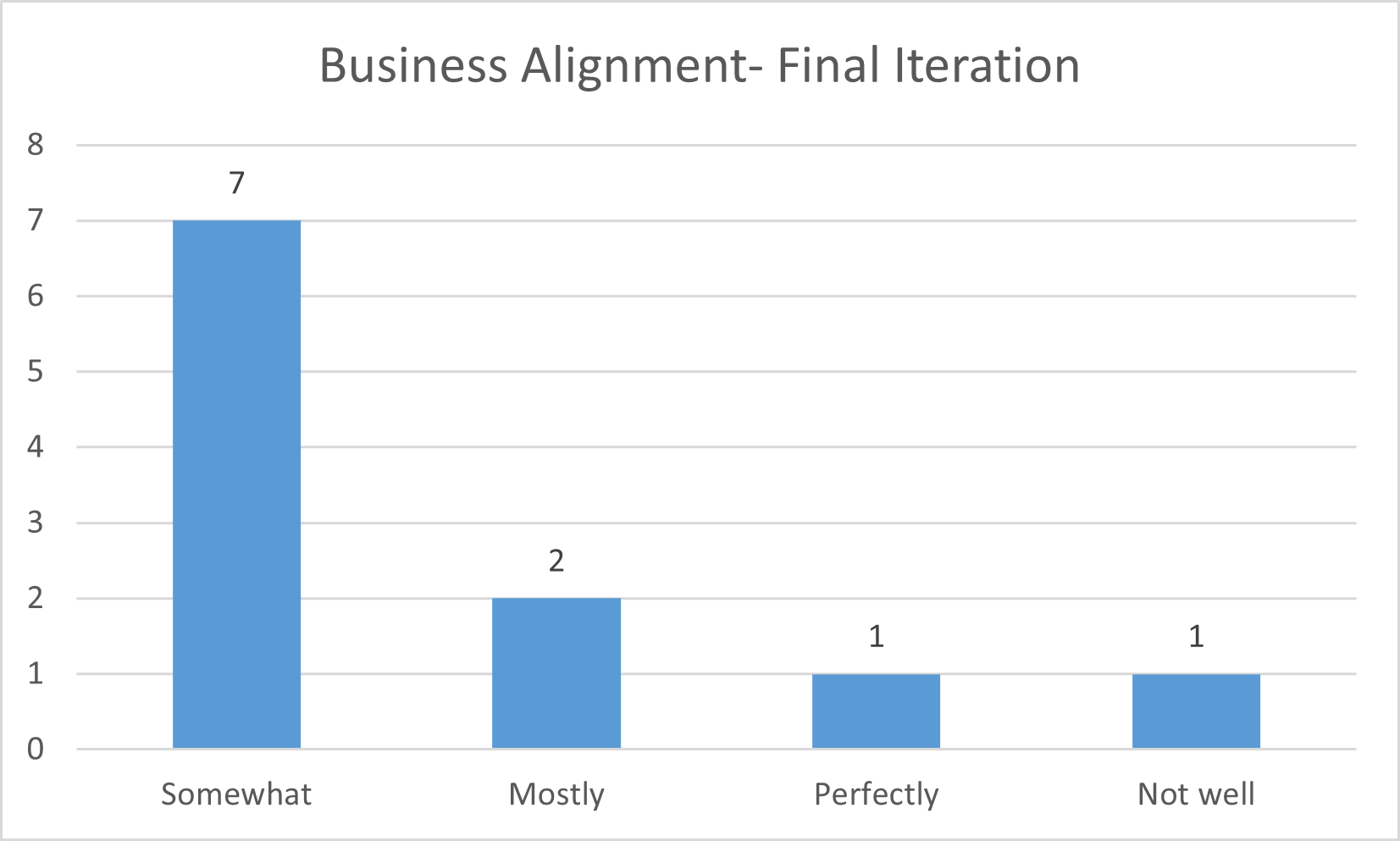}
    \caption{Alignment of system in business need - post updation}
    \label{fig:businessalignment_round2}
\end{figure}

Regarding the usefulness after the augmented knowledge base, participants provided varied responses on how well the chatbot aligned with business needs. Seven participants rated it as "somewhat", two as "mostly", one as "perfectly", one as "not at all", and one as "not well". This distribution indicates a moderate but mixed perception of the relevance of the augmented knowledge base to business needs. Figure \ref{fig:businessalignment_round2} is a bar chart showing the distribution of the ratings regarding usefulness after the augmentation. 

\subsubsection{Summary of Findings}
The evaluation results after updating the knowledge base indicate a moderate but noticeable improvement in the overall performance of the chatbot and practical utility. The integration of synthetic personas and segment-specific information increased the average accuracy rating from 5.88 to 6.42.It can also be concluded that the system was able to handle complex queries, as none of the participants selected the "never" option.

Regarding practical utility, 81.82\% of evaluators rated the augmented knowledge base as at least "somewhat useful", indicating a positive trend in the usefulness of the system in the business contexts. Despite this positive indication, there remains room for further refinement to fully align the system's output with business needs and expectations. Overall, while the augmentation has contributed to improved performance, further refinement can be made to fully optimize the effectiveness and relevance of the chatbot to business needs.

\section{Discussion}
\label{sec:Discussion}

A comprehensive analysis of each of the research questions and the potential reasons for the observed outcomes is discussed in this section.
\begin{enumerate}
    \item \textbf{Effectiveness of Persona-Based Chatbot in Decision-Making}
    
    The evaluation of the persona-based chatbot revealed moderate effectiveness in supporting decision-making and automating customer-facing processes. Even though the system was able to provide relevant responses, the average rating of 5.88/10 indicates that there were inconsistencies in providing accurate and contextually appropriate information. One of the reasons for these inconsistencies could be limited data used. When it comes to ability of the system to reduce workload, 75\% of participants believed that the system had the potential to reduce workload and automate routine tasks.

    \item \textbf{Effectiveness of Prompting Techniques in Synthetic Persona Generation}
    
    From the generated personas it can be observed when multiple customers were mentioned in a single customer success story, the LLM tend to focus on only one customer. The relevant information about others customers in the same story are often missed. This tendency to concentrate on a single person could potentially reduce the completeness of the persona generated. This suggests that further refinement in data processing or prompt engineering may be needed to make sure a comprehensive persona is generated in cases where multiple customer information are present in single stories.
    
    The evaluation of the generated personas was conducted with a small sample size, consisting of only three evaluators and a limited number of personas. This limited scope may affect the generalizability of the findings. Future evaluations with a larger sample could provide better insights. From the statistical test it is evident that the Few-Shot prompting produced more comprehensive personas. However, CoT prompting outperformed Few-Shot in terms of efficiency. This personas generated by this method took less time and  consumed fewer tokens. This finding is particularly relevant in the context of real-world deployment, where response time and resource consumption are critical factors. The lack of significant differences in relevance and consistency between the two methods indicates that both prompting techniques was similar in terms  factual accuracy and reducing fabricated content. This suggests that while Few-Shot may be preferable for completeness, CoT is an optimal choice for scenarios where response efficiency is prioritized.
    
    \item \textbf{Impact of Knowledge Base Augmentation}
    
    The evaluation of the augmented chatbot was conducted with participants from different business functions, including R\&D, customer relations, and IT. Their expectations and query types varied significantly, which may have influenced their assessment. For example, IT engineers might ask for more technical information, while customer experience staff might ask queries related to customer data. Addressing the chatbot functionalities for different users may help to improve overall satisfaction. Augmenting the knowledge base with synthetic personas and segment-specific information resulted in a slight increase in accuracy, with the average rating rising to 6.42. 81.82\% of participants rated the augmented knowledge base as at somewhat useful. This highlights the need for further refinement in data selection and segmentation.
\end{enumerate}

The analysis indicates that while the persona-based chatbot demonstrated some effectiveness in automating customer-facing processes and supporting decision-making, its overall performance was limited by data limitations and variability in response quality. The Few-Shot prompting method produced more complete personas, whereas CoT prompting was more efficient in terms of response time and token usage. Augmenting the knowledge base has shown slight improvements in accuracy and perceived usefulness. Further data refinement and prompt engineering can optimize the system to align more closely with business objectives.

\section{Conclusion, Limitations, and Future Work}

\subsection{Conclusion}
In conclusion, this study demonstrates the potential of leveraging LLMs to generate synthetic customer personas and enhance business decision-making through a persona-based RAG chatbot. By comparing Few-Shot and Chain-of-Thought prompting methods, the research highlights trade-offs between persona completeness and generation efficiency. The integration of synthetic personas into the chatbot's knowledge base led to measurable improvements in response accuracy and user-perceived utility, suggesting that such approaches can effectively complement traditional persona development methods and support scalable, data-driven strategies in industrial settings.

\subsection{Limitations}
The implementation of the conversational system and the persona generation system faced several limitations that impacted the overall scope and performance of the project. 

\begin{enumerate}
    \item \textbf{Knowledge Base}: The knowledge base used for the chatbot was limited to a small set of verified personas and segment information specifically on mining, quarrying, and aggregates. This limits the diversity of the information available to generate responses and potentially reduces the chatbot’s ability to provide comprehensive insights for other relevant segments.
    \item \textbf{Input Data For Synthetic Persona Generation}: For synthetic persona generation, the data source was restricted to customer success stories, which mainly showcased positive customer experiences. These narratives often emphasized successful outcomes rather than describing the challenges faced or areas for improvement. Consequently, the generated personas may lack a balanced perspective, as they primarily reflect favorable customer experiences with VCE products. 
    \item \textbf{Evaluation}: The evaluation process was subjective, as different evaluators might have a different opinion on the quality of the persona. This subjectivity could introduce potential biases and inconsistencies in the assessment results.
\end{enumerate}

\subsection{Future Work}
To make the conversational system and the persona generation better, there are several areas to work on in the future. 
\begin{enumerate}
    \item \textbf{Expand Data}: Currently, data are limited to verified personas and segment-specific information focused on mining, quarrying, and aggregates. Future work could include additional data sources such as customer feedback, satisfaction surveys, and competitor analysis reports. This would provide a more comprehensive dataset that captures diverse customer experiences.
    \item \textbf{Persona Generation}: Another area for further development is the refinement of persona generation techniques. The present approach utilizes GPT-4o Mini with prompting methods. Future research could explore more advanced LLMs or using a fine-tuned LLM with information about VCE. This approach could potentially improve the contextual accuracy and relevance of the personas generated. 
    \item \textbf{Real-Time Updation:} As VCE customer needs and challenges change over time, implementing mechanisms to track changes in customer personas over time would be beneficial. This could be achieved by periodically updating the knowledge base with new customer success stories and feedback data, allowing the system to maintain relevance over time.
    \item \textbf{Chatbot Framework}: When it comes to the chatbot, exploring advanced RAG frameworks such as Graph-RAG \cite{han2024retrieval} and Multi-Hop RAG \cite{tang2024multihop} could improve the system. Graph-RAG introduces graph structures to link related personas, customer success stories, and other data points. This approach could capture more complex relationships between data elements, enabling richer and more context-aware retrieval. Similarly, Multi-Hop RAG \cite{tang2024multihop} extends the RAG framework by retrieving and sequentially processing multiple data points to generate more comprehensive responses. Another promising direction includes improving the chatbot’s interaction to match specific user roles. For example, tailoring the conversation paths for specific roles, such as R\&D Engineers or Customer Experience Teams. This could potentially improve the relevance of responses and provide more targeted insights.
    \item \textbf{Evaluation Metrics:} Lastly, including automated metrics for evaluating the retrieval system and persona quality could streamline the evaluation process and reduce dependence on subjective human evaluation.
\end{enumerate}

\section*{Acknowledgments}
The authors would like to express their sincere gratitude to the Department of Future Solutions, the Department of Brand Experience, the Department of Digital\&IT, and the Department of Finance at Volvo Construction Equipment for their support and guidance throughout the project.

\end{document}